\documentclass[10pt,journal,compsoc]{IEEEtran}
\usepackage{array}
\usepackage{amsmath}
\usepackage{amsfonts}
\usepackage{amssymb}  
\usepackage{graphics} 
\usepackage{graphicx} 
\usepackage{epstopdf}
\usepackage{subfigure}
\usepackage{float}
\usepackage{url}
\usepackage{amsthm}
\usepackage{makecell}
\usepackage{bm}
\usepackage{multirow}
\usepackage{times}

\usepackage{algorithm}
\usepackage{algorithmic}

plus 3pt minus 3pt \dblfloatsep =5pt plus 3pt minus 3pt \floatsep=5pt plus 3pt minus 2pt \abovecaptionskip=5pt plus 2pt minus
2pt \belowcaptionskip=5pt plus 3pt minus 3pt

\ifCLASSOPTIONcompsoc
  \usepackage[nocompress]{cite}
\else
  \usepackage{cite}
\fi
\ifCLASSINFOpdf
\else
\fi

\begin{document}
%
\title{MTFH: A Matrix Tri-Factorization Hashing Framework for Efficient Cross-Modal Retrieval}
%
%
%


\author{Xin~Liu, Zhikai Hu, Haibin Ling, and  Yiu-ming~Cheung,~\IEEEmembership{Fellow, IEEE}
\thanks{This work was supported by the National Science Foundation of China (No. 61673185 and No. 61672444),  Fundamental Research Funds for the Central Universities of Huaqiao University (No. ZQN-PY309), and  the Faculty Research Grant of Hong Kong Baptist University (No. FRG2/17-18/082).

X. Liu is with Department of Computer Science, Huaqiao University,  Xiamen, 361021, China, and also
 with  State Key Laboratory of Integrated
Services Networks, Xidian University, Xi'an, China. E-mail: xliu@hqu.edu.cn

Z.K. Hu is with Department of Computer Science \& Fujian Key Laboratory of Big Data Intelligence and Security, Huaqiao University,  Xiamen, 361021, China. E-mail: zkhu@hqu.edu.cn

H. Ling is with the Department of Computer Science, Stony Brook University, Stony Brook, 11794, USA. E-mail: haibin.ling@stonybrook.edu

Y.M. Cheung is with  Department of Computer Science, Hong Kong Baptist University,  Hong Kong SAR, China. E-mail: ymc@comp.hkbu.edu.hk}
}



\markboth{to appear in IEEE Transactions on Pattern Analysis and Machine Intelligence}%
{Shell \MakeLowercase{\textit{et al.}}: Bare Demo of IEEEtran.cls for Computer Society Journals}
%


\IEEEtitleabstractindextext{%
\begin{abstract}
Hashing has recently sparked a great revolution in cross-modal retrieval
because of its low storage cost and high query speed. Recent cross-modal hashing methods often learn unified or equal-length hash codes  to represent
the multi-modal data and make them intuitively comparable. However, such unified or equal-length hash representations could inherently sacrifice their representation scalability because the data from different modalities may not have one-to-one correspondence and could be encoded more efficiently by different hash codes of unequal lengths. To mitigate these problems, this paper exploits a related and relatively unexplored problem: encode the heterogeneous data with varying hash lengths and generalize the cross-modal retrieval in various challenging scenarios. To this end,  a generalized and flexible cross-modal hashing framework, termed Matrix Tri-Factorization Hashing (MTFH), is proposed to work seamlessly in various settings including paired or unpaired multi-modal data, and equal or varying hash length encoding scenarios. More specifically, MTFH exploits an efficient objective function to flexibly learn the modality-specific hash codes with different length settings, while synchronously learning two semantic correlation matrices to semantically correlate the different hash representations for heterogeneous data comparable. As a result, the derived hash codes are more semantically meaningful for various challenging cross-modal retrieval tasks. Extensive experiments evaluated on public benchmark datasets highlight the superiority of MTFH under various  retrieval scenarios and show its competitive performance with the state-of-the-arts.
\end{abstract}

\begin{IEEEkeywords}
Cross-modal retrieval, matrix tri-factorization hashing, varying hash length, semantic correlation matrix.
\end{IEEEkeywords}}


\maketitle

\IEEEdisplaynontitleabstractindextext

%
\IEEEpeerreviewmaketitle

\IEEEraisesectionheading{\section{Introduction}\label{introduction}}

\IEEEPARstart{W}{ith} the explosive growth of multi-modal data in  social networks, the relevant data from different modalities  often endow semantic correlations, and there is an immediate need for effectively analyzing the data across different modalities. In recent years, cross-modal retrieval, which enables similarity search across heterogeneous modalities, has attracted a great amount of attention in information retrieval community. In the general setting of the problem, a user searches for semantically relevant results of one modality in response to a query item of another different modality, \emph{e.g.}, images that visually illustrate the topic of a textual query, or textual descriptions that concretely describe the contents of a visual query. Nevertheless, the multi-modal data usually span in different feature spaces, and such heterogeneous property has been widely considered as a great challenge
to cross-modal retrieval. In order to eliminate such diversity between different modalities, an intuitive way is  to learn a common latent subspace  so that the mapping features in such subspace can be directly compared~\cite{sharma2012generalized,pereira2014role,peng2016semi}. However, the main drawback of these subspace methods is the level of computational complexity to deal with the large-scale and high dimensional multi-modal data.

\begin{figure}[!t]
	\centering
	\includegraphics[width=9cm]{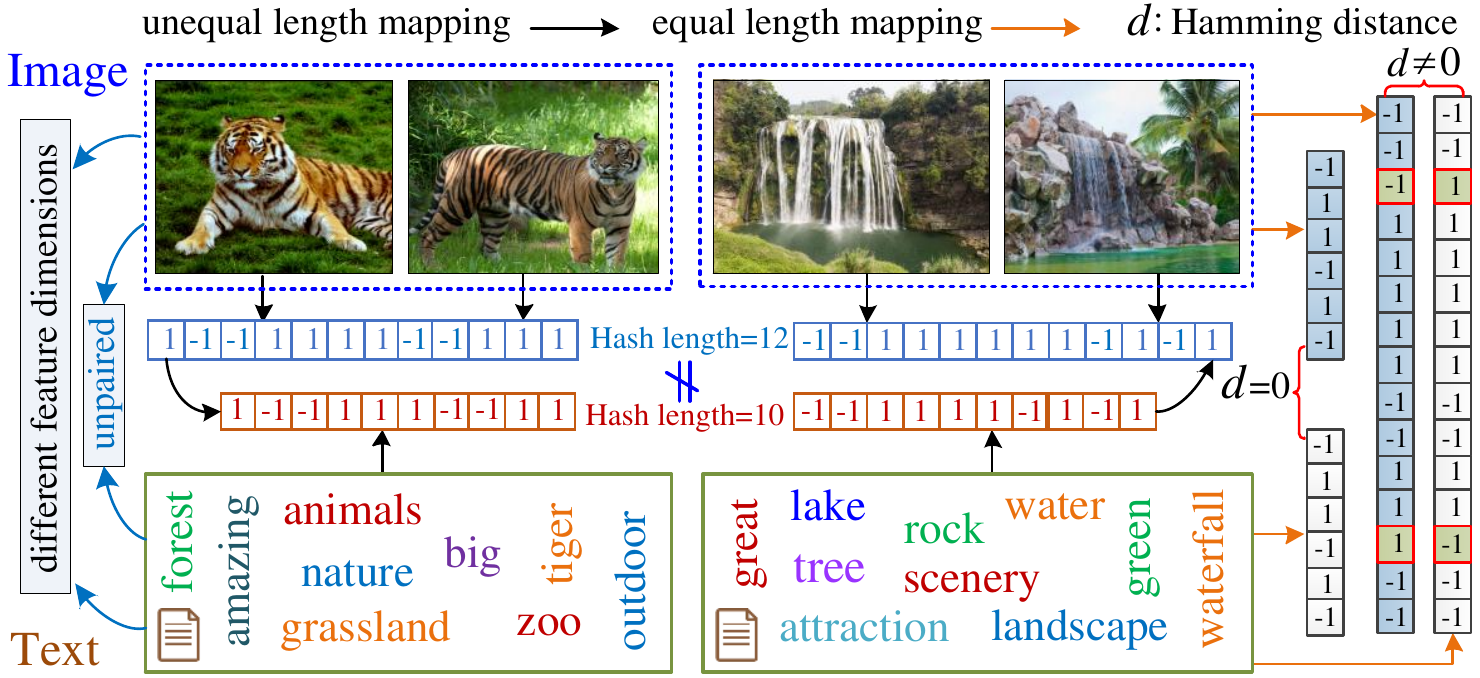}
\vspace{-0.5cm}
	\caption[Introduction]{Two typical examples show that one image may be annotated with multi-labels and one text paragraph may depict multiple relevant images. Meanwhile,
the heterogeneous modalities often have different feature dimensions, and the hash codes of heterogeneous modalities stored in database may have equal or unequal lengths in practice.}
	\label{unpair}
\end{figure}

In recent years, cross-modal hashing~\cite{masci2014multimodal,wu2014sparse}  is gaining significant popularity due to its low storage cost, fast retrieval speed and impressive retrieval performance. It aims to transform the high-dimensional data into compact
binary codes and generate similar binary codes for the relevant samples from different modalities.   Although various kinds of cross-modal hashing attempts have been investigated to correlate the heterogeneous modalities, it remains a challenging task to achieve efficient cross-modal retrieval mainly due to the complex integration of semantic gap, heterogeneity and diversity within the heterogeneous data samples. For instance, the feature representations of  heterogeneous modalities often  have different physical meanings and numerical dimensionalities  with incomparable space structures. Further, as shown in Fig.~\ref{unpair}, the heterogeneous data may be practically paired (\emph{i.e.}, one-to-one correspondence) or unpaired (\emph{e.g.}, a text paragraph depicts multiple images), and the semantics of each sample may be marked as either single label or multiple labels~\cite{song2013inter}. Therefore,  the widespread existence of these complex multi-modal data has significantly increased the demand of more effective cross-modal hashing technologies to tackle these challenging scenarios.

In the literature, the pioneer cross-modal hashing methods~\cite{bronstein2010data,kumar2011learning} select to separate the equal-length hash code learning for different modalities, and these works often  build a weak connection between heterogeneous data samples. To mitigate this problem, the majority of recent cross-modal hashing approaches project the multi-modal data into a common semantic space and utilize a unified hash code to represent the heterogeneous data point, in either supervised fashion where the labels are provided, or unsupervised fashion where the labels are unavailable. Nevertheless, these approaches mainly focus on the paired multi-modal collections, and very little work \cite{mandal2017generalized} has been designed to handle the unpaired multi-modal scenarios. In addition, as shown in Fig.~\ref{unpair}, an even more challenging scenario may arise in cross-modal retrieval, \emph{i.e.}, the hash representations from heterogeneous modalities could be generally encoded and stored by different code lengths in the database, \emph{e.g.}, a text paragraph is discriminatively encoded by 10 bits while an image by 12 bits. This is practically reasonable because the feature dimensions
of heterogeneous modalities often differ significantly, which necessitates the different hash lengthes for better representation. Note that,
the high retrieval performance for many hashing methods empirically depends
on the appropriate selection of code length~\cite{zhen2012co,PLMH2013,zhang2014large}. On the one hand, the big length of hash code is able to reduce the false
collisions (\emph{i.e.}, non-neighbor samples falling into the same bucket) and generally yields high precision. On the other hand,  the long hash representation of a low-dimensional multimedia data significantly increases the sparsity of the Hamming space, which may induce potential noise and result in a low recall rate. The main reason lies in that the collision probability  that two codes of similar instances fall into the same hash bucket decreases exponentially as the code length increases. An example is illustrated in Fig.~\ref{unpair}. It can be found that two short hash codes of semantically similar instances derived from heterogeneous modalities result in zero Hamming distance, while the mappings to long hash length representations induce nonzero Hamming distance. Under such circumstances, the long hash codes may result in low recall performance. Therefore, an inappropriate hash length selection may make it uncompetitive for challenging cross-modal retrieval tasks, \emph{e.g.}, a very low-dimensional text query to retrieve high-dimensional relevant image samples.

\begin{figure}[!t]
	\centering
	\includegraphics[width=8.6cm]{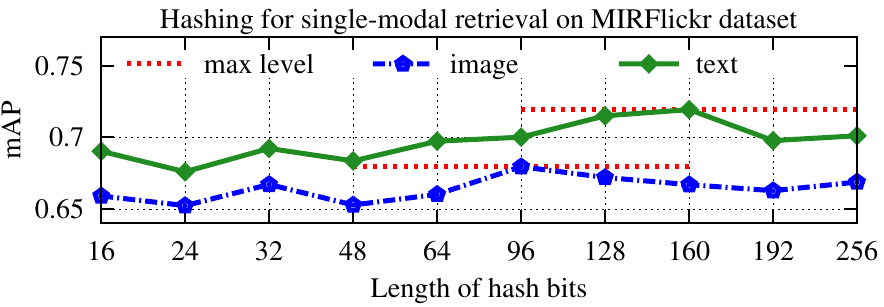}
\vspace{-0.3cm}
	\caption[Introduction2]{Single-modal retrieval results obtained by Fast Supervised Discrete Hashing (FSDH)~\cite{fasthash2018} and tested with different hash lengths.}
	\label{unpair2}
\end{figure}

Remarkably, the representations of multi-modal data in terms of unified or equal-length hash codes are  the common ways to facilitate cross-modal retrieval, and it seems that there is no previous work to surpass such representation assumption. In practice, the code length is of crucial importance to the quality of hash codes because it can be treated as a trade-off  between the discriminative power and redundancy. Suppose that the hash lengths of $q_1$ and $q_2$ (in general $q_1{\neq}q_2$) bits with respect to image and text modalities are optimal for single-modal retrieval, and the best performance can be acquired when the code length reaches an optimal number. Under such circumstances, the hash length setting of $q$ bits ($q{\neq}q_1$ and $q{\neq}q_2$) will naturally bring the negative effect to the retrieval performance. An illustrative example tested on MIRFlickr dataset~\cite{Huiskes2008The} is shown in Fig.~\ref{unpair2},  it can be found that the best retrieval performances are not usually achieved by  large hash codes, and the optimum retrieval results with respect to each modality are not usually produced by the same hash length settings. Therefore, the strictly equalized hash length representation of heterogeneous modalities may inherently sacrifice their representation capability and scalability because it cannot guarantee the learned binary codes to be semantically discriminative for heterogeneous data representation.

%
%
%

In practice, the feature dimensions of heterogeneous modalities may be significantly different, and such physical difference heuristically motivates us to consider different hash lengths for heterogeneous data representations. To the best of our knowledge, varying hash length encoding of multi-modal data and its application to cross-modal retrieval have yet to be explored.  In this paper, we  break the limitations of  equalized hash length representation by allowing varying hash length encoding for different modalities, and seamlessly treat the paired or unpaired multi-modal data collections in an integrated way. To this end,  a generalized and flexible hashing framework, termed \emph{Matrix Tri-Factorization Hashing} (MTFH), is proposed to facilitate various cross-modal retrieval tasks. Specifically, MTFH is a two-stage hashing framework, which allows for less complex formulations in comparison with the coupled formulations.  In  the first stage, MTFH constructs an affinity matrix by semantic label supervision, either square or non-square, depending on the availability of paired or unpaired data samples.  Then, the modality-specific hash codes, of either equal or unequal lengths, are jointly learned with two  semantic correlation matrices. In  the second stage, kernel logistic regression is efficiently utilized to learn the hash mapping functions from feature space to hash code domain. To sum up, the major contributions of this paper are highlighted as follows:

\begin{itemize}
\item A generalized and flexible cross-modal hashing framework is developed, which can work seamlessly in various retrieval tasks including paired or unpaired multi-modal data, and equal or varying hash length encoding scenarios.
\item  MTFH is the first attempt in learning varying hash codes of different lengths for heterogeneous data comparable, and  the learned modality-specific hash codes are more semantically meaningful for cross-modal retrieval.
\item  An efficient discrete optimization algorithm is developed for MTFH without relaxation, which can well reduce the quantization error during the hash code learning process.
\item Extensive experiments on public benchmarks highlight the advantages of MTFH under various cross-modal retrieval tasks and show its comparable or in most cases improved retrieval performance over the existing counterparts..
\end{itemize}

The remainder of this paper is organized as follows. In Section~\ref{relatedworks}, we make an overview of the existing cross-modal hashing works, and in Section~\ref{proposedwork} we elaborate the proposed MTFH framework and its optimization scheme in detail. In Section~\ref{experiments}, we conduct various experiments and comparisons on popular benchmark datasets. Finally, we draw a conclusion  in  Section~\ref{conclusion}.

%

\section{Related Works}\label{relatedworks}
The goal of cross-modal retrieval is to obtain semantically related data samples in one modality for a query in another different modality, and its main difficulty is  to
explicitly measure the content similarity between the heterogeneous samples. Since the heterogeneous data of different modalities often reside in different feature spaces, an
intuitive way is to project the heterogeneous data  into a common subspace and  minimizing their heterogeneities. Along this line, canonical correlation
analysis (CCA)~\cite{hardoon2004canonical}, aiming to learn a latent space that can maximize the correlations between the projected vectors of different modalities,
is popularized for retrieval across different modalities. Accordingly, many reasonable extensions, \emph{e.g.},  bi-linear model (BLM)~\cite{BilinearModels2000}, latent subspace analysis (LSA)~\cite{sharma2012generalized}, sparse subspace learning (SSL)~\cite{wu2014sparse,zhai2014learning}, and correlated subspace learning (CSL)~\cite{pereira2014role,wang2016joint},  have also been developed. Nevertheless, these  methods are generally unsuitable for processing large-scale and high-dimensional multi-modal data.

Hashing technique~\cite{gionis1999similarity,liu2016deep,wang2018survey},  favored for its low storage cost and fast query speed,  has recently attracted much attention in cross-modal retrieval domain.  Most prior hashing works mainly concentrate on producing binary codes for data within the same modality, \emph{e.g.}, locality sensitive hashing (LSH)~\cite{raginsky2009locality} and
its kernelized extension~\cite{kulis2012kernelized}, spectral hashing~\cite{weiss2009spectral} and k-means hashing (KMH)~\cite{he2013k}. These hashing methods provide important theoretical foundations for cross-modal hashing, whose main challenge is to learn the compact binary codes that can construct the underlying correlations between heterogeneous  modalities. In the literature, existing cross-modal hashing methods mainly fall into the modality-independent and modality-dependent branches. Modality-independent approaches primarily
 exploit the separate hash codes and learn the corresponding hash functions for different modalities individually~\cite{bronstein2010data,add2Zhen,kumar2011learning}. For instance,
 cross-view hashing (CVH)~\cite{kumar2011learning} attempts to learn the independent hash codes of different modalities while minimizing the similarity-weighted hamming distances between them.
 Another representative work is multi-modal latent binary embedding (MLBE)~\cite{add2Zhen}, which regards the binary latent factors as hash codes and employs a probabilistic model to learn the  hash functions from multi-modal data independently. However, these methods often build a weak connection between  heterogeneous modalities and their retrieval performances need further improvement.

Modality-dependent approaches mainly learn the unified or correlated hash codes to characterize the multi-modal data, which can be roughly categorized into unsupervised and supervised branches. Without semantic label supervision, unsupervised cross-modal hashing intuitively learns the hash codes from original feature space to Hamming space. For instance, inter-media hashing (IMH)~\cite{song2013inter} first exploits the intra-view and inter-view consistency in a common Hamming space, and then utilizes the linear regression to generate the hash codes. Collective matrix factorization hashing (CMFH)~\cite{ding2016large} employs the joint matrix factorization to learn the unified hash codes for varying multi-modal data, while latent semantic sparse hashing (LSSH)~\cite{zhou2014latent} produces a unified hash code via the latent semantic sparse representation. In addition, fusion similarity hashing (FSH)~\cite{xmu2017} preserves the fusion similarity from multiple modalities and learns the semantically correlated hash codes for heterogeneous data representations. Although these methods are able to capture the semantic correlations between heterogeneous modalities, the hash codes learned in an unsupervised way are not discriminative enough and the corresponding cross-modal similarity is not well preserved in the Hamming space. Consequently, these approaches are restricted by the semantic gap that the high-level semantic hash description of a data sample differs from its low-level feature descriptor, which therefore degrade the retrieval performance.

Supervised cross-modal hashing often utilizes the semantic labels or relevance feedbacks to mitigate the semantic gap between heterogeneous modalities,
which can produce more compact  hash codes to boost the  retrieval performance. Along this line,  semantic correlation maximization (SCM)~\cite{zhang2014large} utilizes the label information to maximize the semantic correlation, while  semantic preserving hashing (SePH)~\cite{lin2015semantics} constructs an affinity matrix by label supervision to  approximate hash codes.  In addition, co-regularized hashing (CRH)~\cite{zhen2012co}, parametric local multi-modal hashing (PLMH)~\cite{PLMH2013}, heterogeneous translated hashing (HTH)~\cite{wei2014scalable}, quantized correlation hashing (QCH)~\cite{wu2015quantized}, supervised matrix factorization hashing (SMFH)~\cite{tang2016supervised} and hetero-manifold regularisation hashing (HMRH)~\cite{zhengPAMI2018}, have also been developed for cross-modal retrieval. It is noted that these supervised methods transform the semantic information of given labels into pairwise similarities and slightly relax the original discrete learning problem into a continuous learning manner, which may yield less effective binary codes due to the accumulated quantization error. To resist such optimization problem, discrete cross-modal hashing (DCH)~\cite{xu2017learning} and cross-modal discrete hashing (CMDH)~\cite{liong2018cross} attempt to directly learn the compact binary codes under a discrete optimization framework. However, these two methods are only designed for the paired multi-modal instances. To adapt unpaired multi-modal data collections, recent generalized semantic preserving hashing (GSePH) ~\cite{mandal2017generalized} factorizes the supervised affinity matrix  to  handle four different cross-modal retrieval scenarios, \emph{i.e.}, single label-paired (SL-P), single label-unpaired (SL-U), multi label-paired (ML-P) and multi label-unpaired (ML-U) scenarios. Similar to most previous works, this method encodes the multi-modal data with equal hash lengths, which may limit its representation discriminability and  scalability in real-world applications, for reason that the data from different modalities  may be practically stored by different hash lengths.

In recent years,  deep neural networks have also been exploited to achieve cross-modal hashing~\cite{cao2016deep,cao2016correlation,jiang2016deep}. Differing from conventional cross-modal hash learning methods, these approaches attempt to combine the high-level feature learning and hash code learning in an integrated way, whereby the feature representations can be optimized with hash code learning through error back-propagation. Although these deep methods have shown outstanding performance on many benchmarks, they are always constrained by computational complexity and exhaustive search for learning  optimum model parameters. Another potential limitation is that these approaches cannot well close the semantic gap between the Hamming distance on binary codes and the metric distance on high-level representations. In addition,  these methods generally utilize the unified hash code to represent the heterogeneous data points and depend highly on paired multi-modal data collections. Therefore, it is still desirable to develop a flexible cross-modal hashing framework practically.

\begin{figure*}[!t]
	\centering
	\includegraphics[width=16cm]{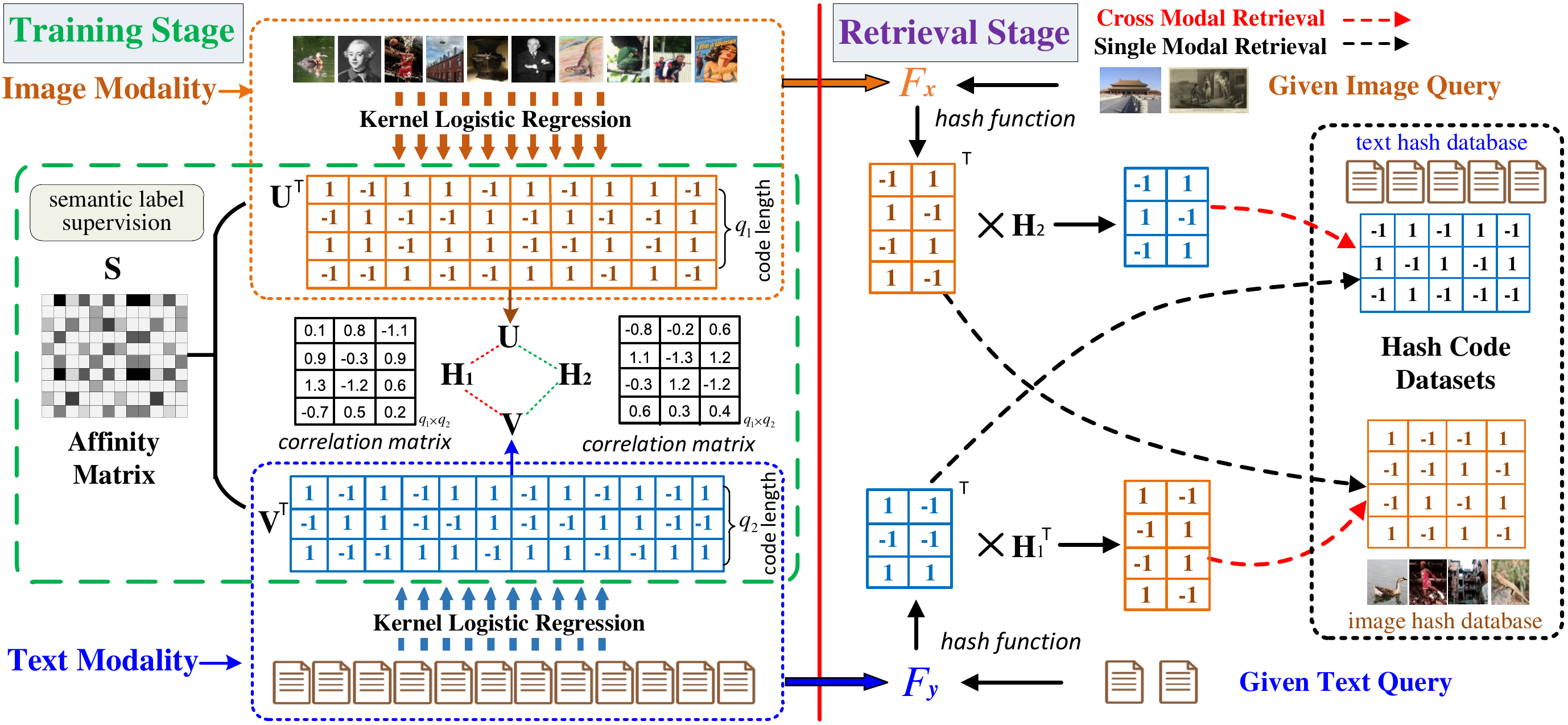}
\vspace{-0.12cm}
	\caption[Introduction]{The proposed  generalized and flexible MTFH framework, which explicitly correlates the  heterogeneous modalities. Note that, MTFH can handle both paired or unpaired multi-modal data collections, and equal or varying hash length encoding  scenarios.}
	\label{framework}
\end{figure*}

\section{Matrix Tri-Factorization Hashing}
\label{proposedwork}

Hashing maps the high-dimensional features into low-dimensional binary codes, while preserving the similarities of data from original space. Although multi-modal relevant data often share the similar semantics, the heterogeneous data samples may not have one-to-one correspondence and their corresponding hash
codes could be practically stored in different lengths. As a typical multi-modal data processing method, matrix factorization~\cite{ding2016large,tang2016supervised} has shown its effectiveness for  cross-modal hashing, but often limits its application domain in unified hash code learning and paired multi-modal data collections. To the best of our knowledge, there has been no previous work on exploring varying hash codes of different lengths for cross-modal retrieval. In this section,  we present an efficient matrix tri-factorization hashing (MTFH) framework to facilitate various kinds of cross-modal retrieval tasks, which can work seamlessly in various settings including paired or unpaired multi-modal data collections, and equal or varying hash length encoding scenarios. To integrate all these challenging tasks, we describe the proposed MTFH framework with only two modalities and its extension problem will be carefully discussed in Section~\ref{extension}. The schematic pipeline of the proposed cross-modal retrieval framework is shown in Fig.~\ref{framework}.

\subsection{Notations and Problem Formulation}
Suppose that we have training data with two different modalities $\mathbf{X}{\in}\mathbb{R}^{n_1\times d_1}$ and $\mathbf{Y}{\in}\mathbb{R}^{n_2\times d_2}$, with $n_1, n_2$ (in some cases $n_1{\neq}n_2$) being the numbers of data samples  and  $d_1, d_2$ (in general $d_1{\neq} d_2$) the feature dimensions of these two modalities, respectively. The provided training labels for  both modalities are $\mathbf{L}_{\mathbf{x}}\in \mathbb{R}^{n_1\times c}$ and $\mathbf{L}_{\mathbf{y}}\in \mathbb{R}^{n_2\times c}$, where $c$ is the number of semantic categories. More specifically, only one of the $c$ entries is equal to 1 if the data is annotated with single semantic label (\emph{e.g.}, $\mathbf{L}^i_{\mathbf{x}}=[0~0~1~0~0]$), and more than
one entries will be equal to 1 if this data is marked with multiple semantic labels (\emph{e.g.}, $\mathbf{L}^j_{\mathbf{y}}=[1~0~1~0~1]$).

 As suggested in~\cite{lin2015semantics}, the semantic affinity matrix with embedding supervision can be efficiently utilized to
learn hash codes of training instances. Accordingly, we first construct an affinity matrix $\mathbf{S}_{ij}{=}\langle\mathbf{L}^i_{\mathbf{x}},\mathbf{L}^j_{\mathbf{y}}\rangle$ or ${\mathbf{S}_{ij}}{=}{e^{{{ - \| {\mathbf{L}^i_{\mathbf{x}} - \mathbf{L}^j_{\mathbf{y}}} \|_2^2} \mathord{\left/ {\vphantom {{ - \| {\mathbf{L}^i_{\mathbf{x}} - \mathbf{L}^j_{\mathbf{y}}} \|_2^2} \sigma }} \right. \kern-\nulldelimiterspace} \sigma }}}$ for both single and multi-label retrieval tasks, where $\langle\cdot,\cdot\rangle$ is the normalized inner product and $\sigma $ a constant factor.
 As demonstrated in~\cite{mandal2017generalized}, an effective hash code learning scheme is to find the optimal hash codes from $\mathbf{S}{\in}\mathbb{R}^{n_1\times n_2}$ directly and attempt to factorize $\mathbf{S}$ as: $\mathbf{S}{\rightarrow}\frac{1}{q_1}\mathbf{U}\mathbf{B}^\mathrm{T}$,  $\mathbf{U}{\in}\mathbb{R}^{n_1\times q_1}$, $\mathbf{B}{\in}\mathbb{R}^{n_2\times q_1}$, where the rows  in $\mathbf{U}$ (resp. $\mathbf{B}$) are the hash codes for the items in $\mathbf{X}$ (resp. $\mathbf{Y}$) and $q_1$ is length of hash code.
 Note that, the values of hash codes are often mapped into $\{-1, 1\}$ for simple computation, and it can be easily mapped into $\{0, 1\}$. It is noted that such a factorization can only generate the  hash codes of equal length for multi-modal instances, which is unsuitable for different hash length encoding scenario.

 Let $q_2$ (in general $q_2{\neq}q_1$) represent another code length  and  $\mathbf{V}{\in}\mathbb{R}^{n_2\times q_2}$ is the targeted hash matrix of $\mathbf{Y}$, it is imperative to learn the correlation between  $\mathbf{B}$ and  $\mathbf{V}$. Since the rows of both $\mathbf{B}$ and  $\mathbf{V}$ characterize the hash codes of the same instance, they share the semantic consistency intrinsically. Therefore, we consider a semantic correlation matrix $\mathbf{H}_1{\in}\mathbb{R}^{q_1\times q_2}$  to map $\mathbf{V}^\mathrm{T}$ into $\mathbf{B}^\mathrm{T}$, \emph{i.e.}, $\mathbf{H}_1\mathbf{V}^\mathrm{T}{\rightarrow}\mathbf{B}^\mathrm{T}$, and propose to factorize $\mathbf{S}$ into three matrices: $\mathbf{S}{\rightarrow}\frac{1}{q_1}\mathbf{U}\mathbf{H}_1\mathbf{V}^\mathrm{T}$. Such a decomposition is a typical matrix tri-factorization (MTF) form~\cite{ding2006orthogonal,xu2017matrix}, and $\mathbf{H}_1$ can map the hash code length of  $\mathbf{Y}$ from $q_2$ to $q_1$, while maintaining the semantic consistency. For cross-modal retrieval with different hash lengths, it is also necessary to map the hash code length of $\mathbf{X}$ from $q_1$ to $q_2$.
 Further, we rewrite $\frac{1}{q_1}\mathbf{U}\mathbf{H}_1\mathbf{V}^\mathrm{T}$ as $\frac{1}{q_2}\mathbf{U}( \frac{q_2}{q_1}\mathbf{H}_1)\mathbf{V}^\mathrm{T}$, and  the length of rows in  $\mathbf{U}(\frac{q_2}{q_1}\mathbf{H}_1)$ becomes $q_2$. That is, $\mathbf{H}_1$ serves as a function of building the semantic connection between two hash representations in the same modality  and ensuring the heterogeneous data comparable between different modalities. Nevertheless, it is infeasible  to derive a single $\mathbf{H}_1$ to  maintain the semantic consistency between different hash representations for both $\mathbf{X}$ and $\mathbf{Y}$. To tackle this problem, we propose to utilize another semantic correlation matrix $\mathbf{H}_2{\in}\mathbb{R}^{q_1\times q_2}$  for the semantic correlation in $\mathbf{X}$, and alternatively factorize $\mathbf{S}$ as: $\mathbf{S}{\rightarrow}\frac{1}{q_2}\mathbf{U}\mathbf{H}_2\mathbf{V}^\mathrm{T}$. It is noted that these triple decompositions have the constraint that the elements of $\mathbf{U}$ and $\mathbf{V}$ take values in $\{-1, 1\}$, and such two factorizations might not exist. To mitigate these problems, we consider the following regularized least squares problem:
	\begin{equation}
		\begin{split}
		\min\limits_{\mathbf{U}{,}\mathbf{V}{,}\mathbf{H}_1{,}\mathbf{H}_2} & \alpha \|\mathbf{S}{-}\frac{1}{q_1}\mathbf{U}\mathbf{H}_1\mathbf{V}^\mathrm{T}\|^2_F {+}(1{-}\alpha)\|\mathbf{S}{-}\frac{1}{q_2}\mathbf{U}\mathbf{H}_2\mathbf{V}^\mathrm{T}\|^2_F \\
		s.t.\;\; & \mathbf{U}{\in}\{-1,1\}^{n_1 \times q_1}, \mathbf{V}{\in}\{-1,1\}^{n_2{\times}q_2} \\
		& \mathbf{V}\mathbf{H}_1^\mathrm{T}{\in}\{-1,1\}^{n_2 \times q_1}, \mathbf{U}\mathbf{H}_2{\in}\{-1,1\}^{n_1{\times}q_2}
		\end{split}
		\label{eq:origin}
	\end{equation}
where $\|\cdot\|_F$ is the Frobenius norm, and $\alpha $ a constant to balance  two learning parts. Remarkably,  the
objective function in Eq.~\eqref{eq:origin} is essentially a challenging combinatorial
optimization problem, which is highly non-convex (usually NP
hard) and cannot be solved trivially by an off-the-shelf solver. Often, a possible solution might involve a deep search of optimal values, which is computationally intractable~\cite{Xia2012Supervised}. Since there are several discrete constraints in Eq.~\eqref{eq:origin},  especially  $ \mathbf{V}\mathbf{H}_1^\mathrm{T}{\in}\{-1,1\}^{n_2 \times q_1}$ and $\mathbf{U}\mathbf{H}_2{\in}\{-1,1\}^{n_1 \times q_2}$, it is impractical to obtain their optimal solutions simultaneously.
To tackle this problem, we introduce two auxiliary variables $\widehat{\mathbf{U}}$ and $\widehat{\mathbf{V}}$ to separate these constraints and
reformulate the Eq. \eqref{eq:origin}  to an approximated one that it can be solved efficiently by employing a regularization algorithm:
	\begin{equation}
		\begin{split}
		\min\limits_{\mathbf{U},\mathbf{V},\widehat{\mathbf{U}},\widehat{\mathbf{V}},\mathbf{H}_1,\mathbf{H}_2} \hspace{-1.5mm}  & \alpha \|\mathbf{S}{-}\frac{1}{q_1}\mathbf{U}\widehat{\mathbf{U}}^\mathrm{T}\|^2_F{+} (1{-}\alpha)\|\mathbf{S}{-}\frac{1}{q_2}\widehat{\mathbf{V}}\mathbf{V}^\mathrm{T}\|^2_F \\
		& + \beta (\|\widehat{\mathbf{U}} - \mathbf{V}\mathbf{H}_1^\mathrm{T}\|^2_F + \|\widehat{\mathbf{V}} - \mathbf{U}\mathbf{H}_2\|^2_F) \\
        & + \lambda (\|\mathbf{H}_1\|^2_F+\|\mathbf{H}_2\|^2_F)\\
		s.t.\;\; & \mathbf{U}\in \{-1,1\}^{n_1 \times q_1}, \mathbf{V}\in \{-1,1\}^{n_2 \times q_2} \\
		& \widehat{\mathbf{U}}\in \{-1,1\}^{n_2 \times q_1}, \widehat{\mathbf{V}}\in \{-1,1\}^{n_1 \times q_2} \\
		\end{split}
		\label{eq:final}
	\end{equation}
where $\beta$ is the penalty parameter and $\lambda$ the regularization parameter. With an appropriate $\beta$, the solution of  Eq.~\eqref{eq:final} is highly close to Eq.~\eqref{eq:origin}. However, the optimization in Eq.~\eqref{eq:final} is still formulated
as a mixed-integer optimization problem, which is still non-convex  and normally intractable due to the discrete constraints on binary codes. In order to simplify the optimization in Eq.~\eqref{eq:final} and obtain a feasible solution,  an  intuitive way is  to replace the constraint set $\{-1,1\}$  with the continuous valued interval  $[-1,1]$ and make the problem computationally tractable. Although this relaxation scheme greatly reduces the hardness of
the optimization by discarding the discrete constraints, the approximated solution may
accumulate large quantization error as the code length increases. Under such circumstances,  the generated binary codes are less effective~\cite{Shen2015Supervised}, which may significantly degrade the  cross-modal retrieval performances. This is mainly because the discrete constraints are not treated adequately during the learning
procedure. As introduced in~\cite{Shen2015Supervised,xu2017learning}, the discrete optimization technique is able to  learn the binary codes directly under discrete constraints, while simultaneously reducing the quantization error. Inspired by these works, we propose an efficient discrete optimization algorithm to solve  Eq.~\eqref{eq:final}, and  alternately minimize the variables  by an iterative framework  until the convergence is reached.

\subsection{Optimization Phases}\label{sec:opt}
The optimization problem in Eq.~\eqref{eq:final} is a mixed binary optimization problem, which is non-convex with respect
to matrix variables $\mathbf{U},\mathbf{V},\widehat{\mathbf{U}},\widehat{\mathbf{V}}$, $\mathbf{H}_1$ and $\mathbf{H}_2$. Remarkably, it is convex with respect to any single variable  while fixing the other ones. Accordingly, an alternating
optimization technique can be adopted to iteratively and
efficiently solve the optimization problem until
 the convergence is reached. In the following, we elaborate the proposed discrete optimization algorithm  in details.

$\mathbf{H}$-\textbf{step:} Learn  $\mathbf{H}_1$ and $\mathbf{H}_2$ by holding $\mathbf{U},\mathbf{V},\widehat{\mathbf{U}}$ and $\widehat{\mathbf{V}}$ fixed, then the sub-optimization problems derived in Eq.~\eqref{eq:final} becomes:
	\begin{equation}
		\begin{split}
		& \min\limits_{\mathbf{H}_1} \beta \|\widehat{\mathbf{U}} - \mathbf{V}\mathbf{H}_1^\mathrm{T}\|^2_F +\lambda \|\mathbf{H}_1\|^2_F, \\
		& \min\limits_{\mathbf{H}_2} \beta \|\widehat{\mathbf{V}} - \mathbf{U}\mathbf{H}_2\|^2_F +\lambda \|\mathbf{H}_2\|^2_F.
		\end{split}
	\end{equation}

Accordingly, $\mathbf{H}_1$ and $\mathbf{H}_2$ can be computed by a regularized linear regression respectively, and their closed-form solutions are:
	\begin{equation}
		\begin{split}
		& \mathbf{H}_1 = \widehat{\mathbf{U}}^\mathrm{T}\mathbf{V}(\mathbf{V}^\mathrm{T}\mathbf{V} + \lambda \beta^{-1}  \mathbf{I})^{-1}, \\
		& \mathbf{H}_2 = (\mathbf{U}^\mathrm{T}\mathbf{U} + \lambda \beta^{-1} \mathbf{I})^{-1}\mathbf{U}^\mathrm{T}\widehat{\mathbf{V}},
		\end{split}
		\label{eq:H}
	\end{equation}
where $\mathbf{I}$ is an identity matrix.
	
$\mathbf{U}$-\textbf{step}: Learn  $\mathbf{U}$ by fixing variables $\mathbf{V},\widehat{\mathbf{U}},\widehat{\mathbf{V}},\mathbf{H}_1,\mathbf{H}_2$,  and the sub-optimization of Eq.~\eqref{eq:final} is further simplified as:
	\begin{equation}
		\begin{split}
		\min\limits_\mathbf{U}\;\; & \alpha \|\mathbf{S} - \frac{1}{q_1}\mathbf{U}\widehat{\mathbf{U}}^\mathrm{T}\|^2_F + \beta \|\widehat{\mathbf{V}} - \mathbf{U}\mathbf{H}_2\|^2_F \\
		s.t.\;\; & \mathbf{U}\in \{-1,1\}^{n_1 \times q_1}.
		\end{split}
		\label{eq:rewrite}
	\end{equation}

The problem in Eq.~\eqref{eq:rewrite} is NP-hard for directly optimizing the
binary code matrix $\mathbf{U}$. As indicated in~\cite{Shen2015Supervised},  a closed-form solution for one
row of $\mathbf{U}$ can be achieved by fixing all the other rows. By expanding each item, we can rewrite Eq.~\eqref{eq:rewrite} as follows:
	\begin{equation}
		\begin{split}
		\min\limits_\mathbf{U}\;\; & \alpha \|\mathbf{S}\|^2_F - \frac{2\alpha}{q_1}\text{Tr}(\mathbf{S}^\mathrm{T}\mathbf{U}\widehat{\mathbf{U}}^\mathrm{T}) + \frac{\alpha}{q^2_1}\|\mathbf{U}\widehat{\mathbf{U}}^\mathrm{T}\|^2_F \\
		& + \beta \|\widehat{\mathbf{V}}\|^2_F - 2\beta \text{Tr}(\widehat{\mathbf{V}}^\mathrm{T}\mathbf{U}\mathbf{H}_2) + \beta \|\mathbf{U}\mathbf{H}_2\|^2_F \\
		s.t.\;\; & \mathbf{U}\in \{-1,1\}^{n_1 \times q_1}
		\end{split}	\label{eq:simple}
	\end{equation}
where $\text{Tr}(\cdot)$ is the trace norm. According to the algebraic operation of the trace, Eq.~\eqref{eq:simple}  can be further simplified as:
	\begin{equation}
		\begin{split}
		\min\limits_\mathbf{U}\;\; &  \frac{\alpha}{q_1^2}\|\mathbf{U}\widehat{\mathbf{U}}^\mathrm{T}\|^2_F + \beta \|\mathbf{U}\mathbf{H}_2\|^2_F - 2\text{Tr}(\mathbf{P}_1\mathbf{U})\\
		s.t.\;\; & \mathbf{U}\in \{-1,1\}^{n_1 \times q_1}
		\end{split}\label{eq:dcc1}
	\end{equation}
where $\mathbf{P}_1{=}\frac{\alpha}{q_1}\mathbf{\widehat{U}}^\mathrm{T}\mathbf{S}^\mathrm{T}{+}\beta \mathbf{H}_2\widehat{\mathbf{V}}^\mathrm{T}$.
Specifically, coordinate descent method has received extensive attention in recent years due to its effectiveness
for solving large-scale optimization problems~\cite{wright2015coordinate}. As suggested in~\cite{Shen2015Supervised,xu2017learning}, we can learn $\mathbf{U}$ bit by bit and the discrete coordinate descent method can be utilized for optimization~\cite{Shen2015Supervised}.  Without loss of generality, let  $\mathbf{u}$ and $\widehat{\mathbf{u}}$ denote the $l$-th column of $\mathbf{U}$ and $\mathbf{\widehat{U}}$, $\mathbf{h}_2$ and $\mathbf{p}_1$ represent the $l$-th row of $\mathbf{H}_2$ and $\mathbf{P}_1$, $\mathbf{U}'$, $\widehat{\mathbf{U}}'$ and $\mathbf{H}'_2$ are the corresponding matrices of $\mathbf{U}$, $\widehat{\mathbf{U}}$ and ${\mathbf{H}_2}$, respectively, excluding $\mathbf{u}$, $\widehat{\mathbf{u}}$ and $\mathbf{h}_2$, we have
	\begin{equation}
		\begin{split}
		\|\mathbf{U}\mathbf{\widehat{U}}^\mathrm{T}\|^2_F & = \text{const}+\|\mathbf{u}\widehat{\mathbf{u}}^\mathrm{T}\|^2 + 2\text{Tr}(\widehat{\mathbf{U}}'\mathbf{U'}^\mathrm{T}\mathbf{u}\widehat{\mathbf{u}}^\mathrm{T}) \;\;\\
		& = \text{const}+ 2\mathbf{\widehat{u}}^\mathrm{T}\widehat{\mathbf{U}}'\mathbf{U'}^\mathrm{T}\mathbf{u} \;\;
		\end{split}
		\label{eq:split1}
	\end{equation}
	\begin{equation}
		\begin{split}
		\|\mathbf{U}\mathbf{H}_2\|^2_F & = \text{const} + \|\mathbf{u}\mathbf{h}_2\|^2 + 2\text{Tr}({\mathbf{H}'_2}^\mathrm{T}\mathbf{U'}^\mathrm{T}\mathbf{u}\mathbf{h}_2) \\
		& = \text{const} + 2\mathbf{h}_2{\mathbf{H}'_2}^\mathrm{T}\mathbf{U'}^\mathrm{T}\mathbf{u}
		\end{split}
		\label{eq:split2}
	\end{equation}
	\begin{equation}
		\;\;\text{Tr}(\mathbf{P}_1\mathbf{U}) = \text{const} + \mathbf{p}_1\mathbf{u},\;\;\;\;\;\;\;\;\;\;\;\;\;\;\;\;\;\;\;\;\;\;\;\;\;\;\;\;\;\;\;\;
		\label{eq:split3}
	\end{equation}
where $\|\mathbf{u}\mathbf{\widehat{u}}^\mathrm{T}\|^2 = \text{Tr}(\mathbf{\widehat{u}}\mathbf{u}^\mathrm{T}\mathbf{u}\mathbf{\widehat{u}}^\mathrm{T}) = n_1\text{Tr}(\mathbf{\widehat{u}}\mathbf{\widehat{u}}^\mathrm{T}) = n_1{\times}n_2 = \text{const}$, $\|\mathbf{u}\mathbf{h}_2\|^2 = \text{Tr}(\mathbf{h}_2^\mathrm{T}\mathbf{u}^\mathrm{T}\mathbf{u}\mathbf{h}_2) = n_1\text{Tr}(\mathbf{h}_2^\mathrm{T}\mathbf{h}_2) = \text{const}$.

By integrating  Eq.~\eqref{eq:split1}, Eq.~\eqref{eq:split2} and Eq.~\eqref{eq:split3} together, we obtain the following optimization  problem:
	\begin{equation}
		\begin{split}
		\min\limits_\mathbf{u}\;\; & \left(\frac{\alpha}{q_1^2}\mathbf{\widehat{u}}^\mathrm{T}\mathbf{\widehat{U}'}\mathbf{U'}^\mathrm{T} + \beta \mathbf{h}_2{\mathbf{H}'_2}^\mathrm{T}\mathbf{U'}^\mathrm{T} - \mathbf{p}_1\right)\mathbf{u} \\
		s.t.\;\; & \mathbf{u} \in \{-1,1\}^{n_1}.
		\end{split}\label{eq:dcc2}
	\end{equation}

Then, the solution of $\mathbf{u}$ can be computed by
	\begin{equation}
		\mathbf{u} = \text{sign}\left(\mathbf{p}_1^\mathrm{T} - \frac{\alpha}{q_1^2}\mathbf{U'}(\mathbf{\widehat{U}'})^\mathrm{T}\mathbf{\widehat{u}} - \beta \mathbf{U'}{\mathbf{H}'_2}\mathbf{h}_2^\mathrm{T}\right).
		\label{eq:U}
	\end{equation}

$\widehat{\mathbf{U}}$-\textbf{step:} Fix $\mathbf{U},\mathbf{V},\widehat{\mathbf{V}},\mathbf{H}_1,\mathbf{H}_2$, and update $\widehat{\mathbf{U}}$, then the sub-optimization problem in Eq. \eqref{eq:final} becomes
	\begin{equation}
		\begin{split}
		\min\limits_{\widehat{\mathbf{U}}}\;\; & \alpha \|\mathbf{S} - \frac{1}{q_1}\mathbf{U}\widehat{\mathbf{U}}^\mathrm{T}\|^2_F + \beta \|\widehat{\mathbf{U}} - \mathbf{V}\mathbf{H}_1^\mathrm{T}\|^2_F \\
		s.t.\;\; & \widehat{\mathbf{U}}\in \{-1,1\}^{n_2 \times q_1}.
		\end{split}\label{hatU}
	\end{equation}

 Similarly, a closed-form solution for one row of $\widehat{\mathbf{U}}$ can be achieved by fixing all the other rows.  By expanding each item, we can rewrite Eq.~\eqref{hatU} as follows:
	\begin{equation}
		\begin{split}
		\min\limits_{\widehat{\mathbf{U}}}\;\; & \alpha \|\mathbf{S}\|^2_F - \frac{2\alpha}{q_1}\text{Tr}(\mathbf{S}^\mathrm{T}\mathbf{U}\widehat{\mathbf{U}}^\mathrm{T}) + \frac{\alpha}{q^2_1}\|\mathbf{U}\widehat{\mathbf{U}}^\mathrm{T}\|^2_F \\
		& + \beta \|\widehat{\mathbf{U}}\|^2_F - 2\beta \text{Tr}(\widehat{\mathbf{U}}^\mathrm{T}\mathbf{V}\mathbf{H}_1^\mathrm{T}) + \beta \|\mathbf{V}\mathbf{H}_1^\mathrm{T}\|^2_F \\
		s.t.\;\; & \widehat{\mathbf{U}}\in \{-1,1\}^{n_2 \times q_1}.
		\end{split}	\label{eq:simple2}
	\end{equation}

Since affinity matrix $\mathbf{S}$ is a fixed item and $\|\widehat{\mathbf{U}}\|^2_F={n_2}{\times}{q_1}=\text{const}$, the above equation can be further simplified as:
	\begin{equation}
		\begin{split}
		\min\limits_{\widehat{\mathbf{U}}}\;\; &  \frac{\alpha}{q_1^2}\|\mathbf{U}\widehat{\mathbf{U}}^\mathrm{T}\|^2_F - 2\text{Tr}(\mathbf{P}_2\widehat{\mathbf{U}})\\
		s.t.\;\; & \widehat{\mathbf{U}}\in \{-1,1\}^{n_2 \times q_1}
		\end{split}\label{eq:dcc3}
	\end{equation}
where $\mathbf{P}_2 = \frac{\alpha}{q_1}\mathbf{U}^\mathrm{T}\mathbf{S}+ \beta \mathbf{H}_1\mathbf{V}^\mathrm{T}$. Let $\mathbf{p}_2$ denote  the  $l$-th row of $\mathbf{P}_2$, we can obtain $\text{Tr}(\mathbf{P}_2\widehat{\mathbf{U}}) = \text{const} + \mathbf{p}_2\mathbf{\widehat{u}}$. According to Eq.~\eqref{eq:split1}, the solution of $\mathbf{\widehat{u}}$ can be achieved by:
	\begin{equation}
		\mathbf{\widehat{u}}= \text{sign}\left(\mathbf{p}_2^\mathrm{T} - \frac{\alpha}{q_1^2}\mathbf{\widehat{U}'}\mathbf{U'}^\mathrm{T}\mathbf{u}\right).
		\label{eq:Uh}
	\end{equation}
	
${\mathbf{V}}$-\textbf{step:}  Learn $\mathbf{V}$ by fixing the variables $\mathbf{U},\widehat{\mathbf{U}},\widehat{\mathbf{V}},\mathbf{H}_1,\mathbf{H}_2$, the sub-optimization problem in Eq.~\eqref{eq:final} can be simplified  as:
	\begin{equation}
		\begin{split}
		\min\limits_\mathbf{V}\;\; & (1 - \alpha) \|\mathbf{S} - \frac{1}{q_2}\widehat{\mathbf{V}}\mathbf{V}^\mathrm{T}\|^2_F + \beta \|\widehat{\mathbf{U}} - \mathbf{V}\mathbf{H}_1^\mathrm{T}\|^2_F \\
		s.t.\;\; & \mathbf{V}\in \{-1,1\}^{n_2 \times q_2}.
		\end{split}\label{Voptimation}
	\end{equation}

Similarly, a closed-form solution for one row of $\mathbf{V}$ can be achieved by fixing all the other rows.
By expanding each item, we can rewrite Eq.~\eqref{Voptimation} as follows:
\begin{equation}
		\begin{split}
		\min\limits_{\mathbf{V}}\; \; & (1-\alpha) \|\mathbf{S}\|^2_F - \frac{2(1-\alpha)}{q_2}\text{Tr}(\mathbf{S}^\mathrm{T}\widehat{\mathbf{V}}\mathbf{V}^\mathrm{T})\\ &+\frac{(1-\alpha)}{q^2_2}\|\widehat{\mathbf{V}}\mathbf{V}^\mathrm{T}\|^2_F + \beta \|\widehat{\mathbf{U}}\|^2_F \\
		& - 2\beta \text{Tr}(\widehat{\mathbf{U}}^\mathrm{T}\mathbf{V}\mathbf{H}_1^\mathrm{T}) + \beta \|\mathbf{V}\mathbf{H}_1^\mathrm{T}\|^2_F \\
		s.t.\;\; & \mathbf{V}\in \{-1,1\}^{n_2 \times q_2}.
		\end{split}	\label{eq:simple3}
	\end{equation}

Since $\mathbf{S}$ and $\widehat{\mathbf{U}}$ are the fixed items, the above equation can be further simplified as:
	\begin{equation}
		\begin{split}
		\min\limits_\mathbf{V}\;\; &  \frac{1-\alpha}{q_2^2}\|\widehat{\mathbf{V}}\mathbf{V}^\mathrm{T}\|^2_F + \beta \|\mathbf{V}\mathbf{H}_1^\mathrm{T}\|^2_F - 2\text{Tr}(\mathbf{P}_\mathbf{3}\mathbf{V})\\
		s.t.\;\; & \mathbf{V}\in \{-1,1\}^{n_2 \times q_2}
		\end{split}\label{eq:dcc4}
	\end{equation}
where $\mathbf{P}_3{=}\frac{1 - \alpha}{q_2}\widehat{\mathbf{V}}^\mathrm{T}\mathbf{S} + \beta \mathbf{H}_1^\mathrm{T}\widehat{\mathbf{U}}^\mathrm{T}$. Without
loss of generality, let  $\mathbf{v}$, $\mathbf{\widehat{v}}$ and $\mathbf{h}_1$  denote the $t$-th column of $\mathbf{V}$, $\widehat{\mathbf{V}}$ and $\mathbf{H}_1$ respectively, $\mathbf{p}_3$ represent the $t$-th row of $\mathbf{P}_3$, $\mathbf{V'}$, $\mathbf{\widehat{V}'}$ and $\mathbf{H}'_1$ are the corresponding matrices of $\mathbf{V}$, $\widehat{\mathbf{V}}$ and ${\mathbf{H}_1}$ respectively excluding $\mathbf{v}$, $\mathbf{\widehat{v}}$ and $\mathbf{h}_1$, we have
the following equations:
\begin{equation}
		\begin{split}
	\;	\|\widehat{\mathbf{V}}\mathbf{V}^\mathrm{T}\|^2_F & = \text{const}+\|\mathbf{\widehat{v}}\mathbf{v}^\mathrm{T}\|^2 + 2\text{Tr}(\mathbf{V'}(\mathbf{\widehat{V}'})^\mathrm{T}\mathbf{\widehat{v}}\mathbf{v}^\mathrm{T})\\
		& = \text{const}+ 2\mathbf{v}^\mathrm{T}\mathbf{V'}(\mathbf{\widehat{V}'})^\mathrm{T}\mathbf{\widehat{v}}
		\end{split}
		\label{eq:Vsplit1}
	\end{equation}
	\begin{equation}
		\begin{split}
		\;\;\|\mathbf{V}\mathbf{H}_1^\mathrm{T}\|^2_F & = \text{const} + \|\mathbf{v}\mathbf{h}_1^\mathrm{T}\|^2 + 2\text{Tr}({\mathbf{H}'_1}\mathbf{V'}^\mathrm{T}\mathbf{v}\mathbf{h}_1^\mathrm{T})\;\;\; \\
		& = \text{const} + 2\mathbf{h}_1^\mathrm{T}{\mathbf{H}'_1}\mathbf{V'}^\mathrm{T}\mathbf{v}\;\;\;
		\end{split}
		\label{eq:Vsplit2}
	\end{equation}
	\begin{equation}
		\text{Tr}(\mathbf{P}_3\mathbf{V}) = \text{const} + \mathbf{p}_3\mathbf{v}.\;\;\;\;\;\;\;\;\;\;\;\;\;\;\;\;\;\;\;\;\;\;\;\;\;\;\;\;\;\;\;\;\;\;\;
		\label{eq:Vsplit3}
	\end{equation}

By integrating the  Eq.~\eqref{eq:Vsplit1}, Eq.~\eqref{eq:Vsplit2} and Eq.~\eqref{eq:Vsplit3} together, we can obtain
the following optimization problem:
	\begin{equation}
		\begin{split}
		\min\limits_\mathbf{v}\;\; & \left(\frac{1-\alpha}{q_2^2}\mathbf{\widehat{v}}^\mathrm{T}\mathbf{\widehat{V}'}\mathbf{V'}^\mathrm{T}+\beta\mathbf{h}_1^\mathrm{T}{\mathbf{H}'_1}\mathbf{V'}^\mathrm{T}-\mathbf{p}_3\right)\mathbf{v} \\
		s.t.\;\; & \mathbf{v} \in \{-1,1\}^{n_1}.
		\end{split}\label{eq:dcc3}
	\end{equation}

Then, the solution of $\mathbf{v}$ can be calculated as:
	\begin{equation}
		\mathbf{v} = \text{sign}\left(\mathbf{p}_3^\mathrm{T} - \frac{1 - \alpha}{q_2^2}\mathbf{V'}(\mathbf{\widehat{V}'})^\mathrm{T}{\mathbf{\widehat{v}}} - \beta \mathbf{V'}{\mathbf{H}'_1}^\mathrm{T}\mathbf{h}_1\right).
		\label{eq:V}
	\end{equation}

$\widehat{\mathbf{V}}$-\textbf{step:}  Fix $\mathbf{U},\mathbf{V},\widehat{\mathbf{U}},\mathbf{H}_1,\mathbf{H}_2$, and update $\widehat{\mathbf{V}}$, then we get the following sub-optimization problem:	
\begin{equation}
		\begin{split}
		\min\limits_{\widehat{\mathbf{V}}}\;\; & (1 - \alpha) \|\mathbf{S} - \frac{1}{q_2}\widehat{\mathbf{V}}\mathbf{V}^\mathrm{T}\|^2_F + \beta \|\widehat{\mathbf{V}} - \mathbf{U}\mathbf{H}_2\|^2_F \\
		s.t.\;\; & \widehat{\mathbf{V}}\in \{-1,1\}^{n_1 \times q_2}.
		\end{split}\label{eq:simple4}
	\end{equation}

By expanding each item, we can rewrite Eq.~\eqref{eq:simple4} as follows:
\begin{equation}
		\begin{split}
		\min\limits_{\widehat{\mathbf{V}}}\; \; & (1-\alpha) \|\mathbf{S}\|^2_F - \frac{2(1-\alpha)}{q_2}\text{Tr}(\mathbf{S}^\mathrm{T}\widehat{\mathbf{V}}\mathbf{V}^\mathrm{T})\\ &+\frac{(1-\alpha)}{q^2_2}\|\widehat{\mathbf{V}}\mathbf{V}^\mathrm{T}\|^2_F + \beta \|\widehat{\mathbf{V}}\|^2_F \\
		& - 2\beta \text{Tr}(\widehat{\mathbf{V}}^\mathrm{T}\mathbf{U}\mathbf{H}_2) + \beta \|\mathbf{U}\mathbf{H}_2\|^2_F  \\
	s.t.\;\; & \widehat{\mathbf{V}}\in \{-1,1\}^{n_1 \times q_2}.
		\end{split}	
	\end{equation}

Since  $\mathbf{S}$ is a fixed item and $\|\widehat{\mathbf{V}}\|^2_F={n_1}{\times}{q_2}=\text{const}$,  the above equation can be further simplified as:
	\begin{equation}
		\begin{split}
		\min\limits_{\widehat{\mathbf{V}}}\;\; &  \frac{1 - \alpha}{q_2^2}\|\widehat{\mathbf{V}}\mathbf{V}^\mathrm{T}\|^2_F - 2\text{Tr}(\mathbf{P}_4\widehat{\mathbf{V}})\\
		s.t.\;\; & \widehat{\mathbf{V}}\in \{-1,1\}^{n_1 \times q_2}
		\end{split}\label{eq:dcc5}
	\end{equation}
where $\mathbf{P}_4{=}\frac{1 - \alpha}{q_2}\mathbf{V}^\mathrm{T}\mathbf{S}^\mathrm{T}{+}\beta \mathbf{H}_2^\mathrm{T}\mathbf{U}^\mathrm{T}$. Let $\mathbf{p}_4$ denote the  $k$-th row of $\mathbf{P}_4$, we can obtain  $\text{Tr}(\mathbf{P}_4\widehat{\mathbf{V}}) = \text{const} + \mathbf{p}_4\mathbf{\widehat{v}}$.
By integrating Eq.~\eqref{eq:Vsplit1}, the solution of $\mathbf{\widehat{v}}$ can be computed by:
	\begin{equation}
		\mathbf{\widehat{v}} = \text{sign}\left(\mathbf{p}_4^\mathrm{T} - \frac{1 - \alpha}{q_2^2}\mathbf{v}{\mathbf{V}'}^\mathrm{T}\widehat{\mathbf{V}}'\right).
		\label{eq:V_}
	\end{equation}	

Accordingly, the optimum elements in Eq.~\eqref{eq:final} can be obtained iteratively  via alternating minimization techniques.

\begin{algorithm}[!htp]
\caption{The Proposed E-RCD for Hash Code Updating}
\textbf{input:} hash matrix $\mathbf{B}\in \{1,-1\}^{n\times q}$, ensemble round $r$; \\
\textbf{output:} updated hash  matrix $\mathbf{\hat B}$; \\
	1: denote  $\mathbf{b}_l$ as the $l$-th column of $\mathbf{B}$;\\
	2: \textbf{for}\; $\tau=1:r$ \;\textbf{do} \\
	3: \;\;\ independent selection at each iteration; \\
	4: \;\;\textbf{repeat} \\
	5:\;\;\;\;choose index $l$ with uniform probability from $\{1,\cdots,q\}$;\\
	6:\;\;\;\;update $\mathbf{b}_{l}^\tau$ via discrete hash learning; \\
	7: \;\;\textbf{until} (all columns are updated ) \\
	8: \textbf{end for} \\
    9: \textbf{return} $\mathbf{\hat B} = \text{sign}\{\mathbf{B}^1+\mathbf{B}^2+\cdots+\mathbf{B}^r\}$.
\label{ERCD}
\end{algorithm}

\subsection{Updating Scheme}
During the coordinate descent optimization, only one variable
is updated at each iteration, while all the others remain fixed. There are several strategies to select the coordinate index, including
cyclic coordinate descent (CCD), randomized  coordinate descent (RCD) and greedy coordinate descent (GCD)~\cite{lin2014accelerated}. More specifically,
CCD updates variables in a cyclic order, while RCD chooses variables randomly based on some
distribution. Differently, GCD measures the coordinate index  by the magnitude of gradient. Since the optimization in our framework is a discrete optimization problem, GCD scheme is improper for this case. In~\cite{Shen2015Supervised,xu2017learning}, discrete cyclic coordinate (DCC) descent scheme is selected to update the binary hash codes. Remarkably, DCC is still an approximated solution to discrete hashing and may
fall into a local minima~\cite{wright2015coordinate,xu2017learning}. To alleviate the possible trapping in local minimum, a straightforward way  is to repeat
the optimization procedures several times with different random initializations. As discussed in~\cite{lin2014accelerated}, empirical studies have proved that RCD locally converges to the global minimum at a geometric rate with high
probability. Specifically, we utilize the ensemble RCD (E-RCD) to derive the hash codes more reliably.

Let $\mathbf{B}{\in}\{1,-1\}^{n\times q}$ be the representative symbol of updating hash code matrix, where $n$ is the number of learning samples and $q$ is the code length. Accordingly, the optimization procedure of the proposed E-RCD is explicitly summarized in Algorithm~\ref{ERCD}.
  Please note that a large  number of rounds in ensemble learning could increase the computational load during the updating process. Fortunately, it is practically adequate to run only a few rounds (\emph{e.g.}, $r{=}3$) in ensemble updating process. Consequently, each elements in Eq.~\eqref{eq:final} can be obtained iteratively by
 repeating each updating process until the procedure converges  or reaches maximum iterations. The main procedures of the proposed MTFH approach are summarized in Algorithm~\ref{algorithm}.

\begin{algorithm}[!htp]
\caption{Matrix Tri-Factorization Hashing (MTFH)}
\textbf{input:} $\mathbf{S}\in \{1,0\}^{n_1\times n_2}$, $q_1$, $q_2$, parameters $\alpha,\beta$; \\
\textbf{output:} $\mathbf{U},\mathbf{V},\mathbf{H}_1,\mathbf{H}_2$; \\
	1: initialize $\mathbf{H}_1,\mathbf{H}_2$ as random matrices, and $\mathbf{U},\mathbf{V},\widehat{\mathbf{U}},\widehat{\mathbf{V}}$ as binary random matrices with elements in $\{-1,1\}$;\\
	2: \textbf{repeat} \\
	3:\;\;\;\;update $\mathbf{H}_1,\mathbf{H}_2$ via Eq.~\eqref{eq:H}; \\
	4:\;\;\;\;compute $\mathbf{u}$ via Eq.~\eqref{eq:U}, update $\mathbf{U}$ via Algorithm~\ref{ERCD}; \\
	5:\;\;\;\;compute $\mathbf{\widehat{u}}$ via Eq.~\eqref{eq:Uh}, update $\widehat{\mathbf{U}}$ via Algorithm~\ref{ERCD}; \\
	6:\;\;\;\;compute $\mathbf{v}$ via Eq.~\eqref{eq:V}, update $\mathbf{V}$ via Algorithm~\ref{ERCD}; \\
	7:\;\;\;\;compute $\mathbf{\widehat{v}}$ via Eq.~\eqref{eq:V_}, update $\widehat{\mathbf{V}}$ via Algorithm~\ref{ERCD}; \\
	8: \textbf{until} (convergency or reaching maximum iterations) \\
    9: \textbf{return} $\mathbf{U},\mathbf{V},\mathbf{H}_1,\mathbf{H}_2$.
\label{algorithm}
\end{algorithm}

\subsection{Learning Hash Functions}
The hash function builds the mapping relation from input
features of each modality to binary codes~\cite{add3chen}.
In general, learning hash functions for any bit of the hash code can be
transformed into a predictive model learning process, and any binary classifier such as linear projections or non-linear projections can be selected to learn the hash function. In the literature, many different hash functions are explored and the most common
hash function is the linear hash function, which projects the
input feature vector by a linear transformation followed by an
element-wise sign operation. Although linear hash function is
very simple to use, it cannot capture the nonlinearity embedded
in real-world data.  To handle non-linear mapping,
kernel logistic regression,  capable of modelling non-linear mappings, is popularized to learn the projections from features to hash codes~\cite{bronstein2010data,lin2015semantics}. For simplicity, we select modality $\mathbf{X}$ for illustration. That is, a non-linear function $\phi$ first  maps the sample $\mathbf{x}_i$ into the reproducing kernel Hilbert space (RKHS) as $\phi(\mathbf{x}_i)$, and then a linear function $f$ in the RKHS space brings the input to the hash code domain. To learn such projection in RKHS for the $k$-th bit ($1\leq{k}{\leq}q_1$), we need to learn the  projection $f^{(k)}_\mathbf{x}$ by minimizing the following function:
\begin{equation}
\mathop {\min }\limits_{{f^{(k)}_\mathbf{x}}} \sum\limits_{i = 1}^{n_1} {\log (1 + {e^{ - \mathbf{b}_i^{(k)}\phi ({\mathbf{x}_i}){f^{(k)}_\mathbf{x}}}})}  + \eta\left\| {{f^{(k)}_\mathbf{x}}} \right\|_2^2
\end{equation}
where $\mathbf{b}_i^{(k)}{\in}\{-1,1\}$ is the $i$-th entry in $\mathbf{b}^{(k)}$, and  $\eta$ is a parameter for weighting the regularizer. For features coming from modality $\mathbf{X}$, we can learn a set of hash functions $F_\mathbf{X}{=}\{f_\mathbf{x}^{(1)},f_\mathbf{x}^{(2)},...,f_\mathbf{x}^{(q_1)}\}$. Similarly, we can also learn a set of hash functions $F_\mathbf{Y}{=}\{f_\mathbf{y}^{(1)},f_\mathbf{y}^{(2)},...,f_\mathbf{y}^{(q_2)}\}$ to map the features  from $\mathbf{Y}$ to the hash code domain. For the testing data $\mathbf{x}$ and $\mathbf{y}$ coming respectively from $\mathbf{X}$ and $\mathbf{Y}$ modalities, the hash codes can be computed as:  $ h_\mathbf{x}{=}\text{sign}(F_\mathbf{X}(\mathbf{x}))$ and $h_\mathbf{y}{=}\text{sign} (F_\mathbf{Y}(\mathbf{y}))$.

\subsection{Hash Codes  for Out-of-Sample Extension}

For any data point  not in the training set, we can
predict its hash code with the corresponding
probability obtained from kernel logistic regression.
For instance, given an unseen instance $\mathbf{x}$ from the modality $\mathbf{X}$, the corresponding output probability for the $k$-th bit of its predicted hash code $h_{\mathbf{x}}^l$ can be calculated as:
\begin{equation}\label{kernelfuntion}
{{\Pr(h_\mathbf{x}^k= b}}| \mathbf{x}) = {\left( {1 + {e^{ - b\phi (\mathbf{x})f_\mathbf{x}^{(k)}}}} \right)^{ - 1}}
\end{equation}
where $b{\in}\{-1, 1\}$ denotes the binary state in hash code and $f_\mathbf{x}^{(k)}$ is the $k$-th projection function in kernel logistic regression. Accordingly, for unseen instances, $\mathbf{x}$ and $\mathbf{y}$, respectively, from modalities $\mathbf{X}$ and $\mathbf{Y}$, we can get their corresponding hash codes $h_{\mathbf{x}}^k$ at the $k$-th bit and $h_{\mathbf{y}}^t$ at the $t$-th bit as follows:
	\begin{equation}
	\begin{split}
		& h_{\mathbf{x}}^k = \text{sign}(\Pr(h_{\mathbf{x}}^k=1|\mathbf{x})-\Pr(h_{\mathbf{x}}^k=-1|\mathbf{x})) \\
		& h_{\mathbf{y}}^t = \text{sign}(\Pr(h_{\mathbf{y}}^t=1|\mathbf{y})-\Pr(h_{\mathbf{y}}^t=-1|\mathbf{y})).
	\end{split}
	\end{equation}

These two modality-specific hash codes are learned independently for single-modal retrieval, and their hash lengths may be different. Fortunately,  with semantic correlation matrices $\mathbf{H}_1$ and $\mathbf{H}_2$, these hash codes can be further transformed into the semantically equivalent patterns to adapt to cross-modal retrieval:
	\begin{equation}
		\widehat{h}_{\mathbf{x}} = \text{sign}(h_{\mathbf{x}}\mathbf{H}_2), \;\;\widehat{h}_{\mathbf{y}} = \text{sign}(h_{\mathbf{y}}\mathbf{H}_1^\mathrm{T}).
	\end{equation}

\subsection{Complexity Analysis}

The computational complexity of the proposed MTFH framework mainly involves the optimization in the training phase. The time complexity of each iteration consists of updating $\{\mathbf{H}_1,\mathbf{H}_2\}$, $\mathbf{U}$, $\widehat{\mathbf{U}}$, $\mathbf{V}$ and $\widehat{\mathbf{V}}$, which respectively, involves the computational complexity of $\mathcal{O}(q^2n{+}q^3)$, $\mathcal{O}((q^2n^2{+}q^3n)r)$, $\mathcal{O}(q^2n^2r)$, $\mathcal{O}((q^2n^2{+}q^3n)r)$ and $\mathcal{O}(q^2n^2r)$, where $n{=}\max(n_1,n_2)$, $q{=}\max(q_1,q_2)$ and  $r$ is ensemble round. Therefore, the overall complexity is approximated as $\mathcal{O}((rq^2n^2{+}(rq^3{+}q^2)n{+}q^3)t)$, where $t$ is the number of iterations to convergence and it is usually less than 20 in practice. In most experiments, the final solution does not substantially change if we utilize a large round number, and therefore it is appropriate to set the ensemble round $r$  at a very small value (\emph{e.g.}, $r{=}3$).  Therefore, the proposed discrete optimization scheme is suitable for practical cross-modal hashing tasks, and more discussions concerning to the large-scale data processing will be included in Section~\ref{extension}.

\section{Experiments}
\label{experiments}

In this section, we conduct a series of quantitative experiments on public  benchmarks and validate the effectiveness of the proposed approach on  various challenging retrieval tasks.  The source code is made publicly available at: \emph{https://github.com/starxliu/MTFH}.


\subsection{Datasets and Evaluation Protocol}
In the experiments, three popular multi-modal datasets, \emph{i.e.}, Wiki\footnote{http://www.svcl.ucsd.edu/projects/crossmodal/},  MIRFlickr\footnote{http://press.liacs.nl/mirflickr/}  and
NUS-WIDE\footnote{http://lms.comp.nus.edu.sg/research/NUS-WIDE.htm}, are selected for testing, and the main description of each dataset is briefly described as follows:

\textbf{Wiki dataset} consists of 10 categories and 2,866 image-text pairs from the public Wikipedia articles~\cite{pereira2014role}.  Specifically, the image is described by a 128-dimensional SIFT feature vector, while the text  article  is characterized by a 10-dimensional feature vector that is computed by the Latent Dirichlet Allocation (LDA) model. The whole Wiki dataset is split into a training set of 2,173 instances and a testing set of 693 instances.

\textbf{MIRFlickr dataset} comprises 25,000 image-text pairs collected from the popular Flickr website~\cite{Huiskes2008The}, where the images are annotated with textual tags. Specifically, each image is described by a 150-dimensional edge histogram descriptor, while the text is represented by a 500-dimensional feature vector  derived from its binary tagging vectors. Each image-text pair is annotated with one or more of 24 semantic labels. As suggested in \cite{lin2015semantics}, we remove the instances whose textual tags appear less than 20 times or label is not annotated, and take out 5\% of the dataset as the query set and the remaining parts as the training set.

\textbf{NUS-WIDE dataset} includes 269,548 image-text pairs with 81 manually
annotated concepts in total~\cite{Chua2009NUS}. Specifically, each image is represented by a 500-dimensional SIFT feature vector, while each text is described by a 1000-dimensional bag-of-words (BoW) vector. Since some of the labels are scarce and a large part of concepts contain little samples, 186,577 annotated instances are selected from the top 10 most frequent concepts to guarantee that each concept has abundant training samples (abbreviated as \textbf{NUS-WIDE-All}). As NUS-WIDE-all is a larger dataset, it is generally impossible to learn the hash functions on the whole database. Therefore, we  randomly select 100,000 labeled image-text pairs from NUS-WIDE-all database to construct a small dataset (abbreviated as \textbf{NUS-WIDE-100k}), with 5\% pairs as the query set and the remaining parts as the training set. For NUS-WIDE-all dataset, we keep the training samples and testing samples as the same as the selection in NUS-WIDE-100k, and utilize the learned hash functions to generate the hash codes of remaining samples.

The quantitative performance is  evaluated by the popular mean Average
Precision (mAP) over all queries in the query set~\cite{lin2015semantics}: $\frac{1}{{n_q}}\sum\nolimits_{i= 1}^{n_q} {\frac{1}{{{m_i}}}\sum\nolimits_{k = 1}^{{m_i}} {p(k)\delta (k)} }$, where $n_q$ is the sample size of  query set, $m_i$ is the number of ground-truth neighbors relevant to query $i$ in the database, $p(k)$ denotes the precision of top $k$ retrieved results, and $\delta (k){=}1$ if the $k$-th retrieved sample is relevant,
otherwise $\delta (k){=}0$. Given a query of one modality, the goal of each cross-modal
task is to find the relevant neighbors from the database of another modality. That is, the relevant instances corresponding to a given query
are defined as those share as least one  semantic label with the query. The larger mAP  generally indicates the better retrieval performance. We take  the testing set of one modality  as the query set to retrieve the relevant data of another modality, including retrieving text with given image (I$\rightarrow$T) and retrieving image with given text (T$\rightarrow$I). In the experiments, we fix  $\alpha{=}0.5$,  $\lambda{=}0.1$ and $\beta{=}0.1$.

\begin{table*}[t]
\caption{Quantitative comparisons of cross-modal retrieval performance (mAP) on different datasets, and the best results are highlighted in bold.}\setlength{\tabcolsep}{0.09cm}
\vspace{-0.3cm}
\begin{center}
\begin{tabular}{|c|c |cccc |cccc |cccc|cccc|}
\hline
	\multirow{2}{*}{Task} & \multirow{2}{*}{Method}	& \multicolumn{4}{c|}{Wiki} & \multicolumn{4}{c|}{MIRFlickr} & \multicolumn{4}{c|}{NUS-WIDE-100k} & \multicolumn{4}{c|}{NUS-WIDE-All}\cr \cline{3-18}
		& & 16 & 32 & 64 & 128 & 16 & 32 & 64 & 128 & 16 & 32 & 64 & 128 & 16 & 32 & 64 & 128 \cr
	\hline \hline
	\multirow{12}{*}{I$\rightarrow$T}
		 & CMFH~\cite{ding2016large} & 0.2172 & 0.2231 & 0.2316 & 0.2395 & 0.5683 & 0.5684 & 0.5687 & 0.5693 & 0.3428 & 0.3434 & 0.3433 & 0.3432 & 0.3658 & 0.3689 & 0.3689 & 0.3681 \cr
		  & SMFH~\cite{tang2016supervised} & 0.2698 & 0.2900 & 0.2929 & 0.3009 & 0.5913 & 0.5997 & 0.5956 & 0.5986 & 0.3612 & 0.3613 & 0.3628 & 0.3635 & 0.3668 & 0.3678 & 0.3690 & 0.3692\cr
	 & FSH~\cite{xmu2017} & 0.2235 & 0.2316 & 0.2408 & 0.2474 & 0.5893 & 0.6027 & 0.6006 & 0.6022 & 0.4927 & 0.4986 & 0.5015 & 0.5057 & 0.4930 & 0.5000 & 0.5093 & 0.5133 \cr
     & SCM\_orth~\cite{zhang2014large} & 0.1561 & 0.1416 & 0.1336 & 0.1339 & 0.5899 & 0.5800 & 0.5738 & 0.5689 & 0.3990 & 0.3813 & 0.3666 & 0.3572 & 0.3975 & 0.3787 & 0.3665 & 0.3559\cr
		 & SCM\_seq~\cite{zhang2014large} & 0.2341 & 0.2410 & 0.2445 & 0.2569 & 0.6280 & 0.6345 & 0.6385 & 0.6490 & 0.5275 & 0.5414 & 0.5481 & 0.5498 & 0.5266 & 0.5378 & 0.5406 & 0.5436\cr
	  & SePH\_rnd~\cite{lin2015semantics} & 0.2702 & 0.3013 & 0.3135 & 0.3181 & 0.6727 & 0.6804 & 0.6799 & 0.6857 & 0.5347 & 0.5472 & 0.5533 & 0.5574 & 0.5264 & 0.5389 & 0.5539 &0.5527\cr
		 & SePH\_km~\cite{lin2015semantics} & 0.2770 & 0.2964 & 0.3153 & 0.3138 & 0.6736 & 0.6789 & 0.6822 & 0.6851 & 0.5381 & 0.5517 & 0.5556 & 0.5654 & 0.5357 & 0.5526 & 0.5681 &0.5724\cr
		 & GSePH\_rnd \cite{mandal2017generalized} & 0.2690 & 0.2906 & 0.3101 & 0.3001 & 0.6544 & 0.6664 & 0.6768 & 0.6842 & 0.5194 & 0.5399 & 0.5489 & 0.5699 & 0.4997 & 0.5436 & 0.5428 &0.5496\cr
		 & GSePH\_km \cite{mandal2017generalized} & 0.2778 & 0.2882 & 0.3044 & 0.3040 & 0.6460 & 0.6649 & 0.6725 & 0.6835 & 0.5018 & 0.5370 & 0.5595 & 0.5715 & 0.5006 & 0.5408 & 0.5571 & 	0.5590 \cr
         & DCH~\cite{xu2017learning}  & 0.3410 & \textbf{0.3692} & \textbf{0.3710} & \textbf{0.3783} & 0.6777 & 0.6730 & 0.6883 & 0.6885 & 0.5706 & 0.5939 & 0.5982 & 0.6072 & 0.5108  &	0.5383 & 0.5480 & 0.5501 \cr
           & SRLCH~\cite{DASSF2018}   & 0.3268 & 0.3345 & 0.3225 & 0.3381 & 0.6166 & 0.5924 & 0.6526 & 0.6327 & 0.4362 & 0.4572 & 0.4506 & 0.4612 & 0.3478 & 0.3517 & 0.3513 & 0.3582\cr
		 \cline{2-18}
		 & MTFH\_rnd & 0.3260 & 0.3523 & 0.3454 & 0.3388 & \textbf{0.7515} & 0.7568 & 0.7592 & 0.7636 & 0.6507 & 0.6557 & 0.6744 & 0.6741 &  0.5949 & 0.6144 & 0.6243 & 0.6228  \cr
		 & MTFH\_km & \textbf{0.3413} & 0.3533 & 0.3511 & 0.3349 & 0.7471 & \textbf{0.7606} & \textbf{0.7651} & \textbf{0.7676} & \textbf{0.6554} & \textbf{0.6591} & \textbf{0.6759} & \textbf{0.6751} & \textbf{0.6021} & \textbf{0.6184} & \textbf{0.6282} & \textbf{0.6271} \cr
\hline \hline
	\multirow{12}{*}{T$\rightarrow$I}
		 & CMFH~\cite{ding2016large} &  0.4902 & 0.5077 & 0.5173 & 0.5348 & 0.5646 & 0.5652 & 0.5649 & 0.5653 & 0.3464 & 0.3472 & 0.3473 & 0.3474 & 0.3687 & 0.3698 & 0.3692 & 0.3698
 \cr
		  & SMFH~\cite{tang2016supervised} & 0.6085 & 0.6274 & 0.6308 & 0.6445 & 0.5890 & 0.5909 & 0.5915 & 0.5954 & 0.3524 & 0.3524 & 0.3529 & 0.3538 & 0.3587 & 0.3593 & 0.3606 & 0.3605
 \cr
		  & FSH~\cite{xmu2017} & 0.4805 & 0.4804 & 0.5127 & 0.5182 & 0.5865 & 0.5970 & 0.5965 & 0.5969 & 0.4751 & 0.4785 & 0.4822 & 0.4879 & 0.4729 & 0.4807 & 0.4883 & 0.4909
 \cr
        & SCM\_orth~\cite{zhang2014large} & 0.1521 & 0.1330 & 0.1258 & 0.1207 & 0.5893 & 0.5802 & 0.5719 & 0.5661 & 0.3873 & 0.3714 & 0.3602 & 0.3574 & 0.3883 & 0.3699 & 0.3589 & 0.3546
 \cr
		 & SCM\_seq~\cite{zhang2014large} & 0.2257 & 0.2459 & 0.2494 & 0.2535 & 0.6176 & 0.6234 & 0.6285 & 0.6369 & 0.4952 & 0.5076 & 0.5157 & 0.5174 & 0.4956 & 0.5031 & 0.5124 & 0.5104
 \cr
		& SePH\_rnd~\cite{lin2015semantics} & 0.6428 & 0.6493 & 0.6570 & 0.6672 & 0.7252 & 0.7306 & 0.7374 & 0.7397 & 0.6231 & 0.6491 & 0.6577 & 0.6654 & 0.6103 & 0.6360 & 0.6507 & 0.6487
\cr
		 & SePH\_km~\cite{lin2015semantics} & 0.6402 & 0.6543 & 0.6585 & 0.6674 & 0.7313 & 0.7320 & 0.7381 & 0.7442 & 0.6310 & 0.6546 & 0.6628 & 0.6702 & 0.6143 & 0.6428 & 0.6533 & 0.6649
 \cr
		 & GSePH\_rnd \cite{mandal2017generalized} & 0.6478 & 0.6644 & 0.6679 & 0.6762 & 0.6894 & 0.7046 & 0.7313 & 0.7367 & 0.5871 & 0.6234 & 0.6419 & 0.6638 & 0.5720 & 0.6334 & 0.6308  & 	0.6442 \cr
		 & GSePH\_km \cite{mandal2017generalized} & 0.6445 & 0.6639 & 0.6683 & 0.6755 & 0.6663 & 0.7113 & 0.7269 & 0.7441 & 0.5595 & 0.6379 & 0.6593 & 0.6764 & 0.5780 & 0.6289 & 0.6482 & 	0.6550  \cr
	    & DCH~\cite{xu2017learning}  & 0.6980 & 0.7160 & 0.7172 & 0.7195 & 0.7455 & 0.7559 & 0.7825 & 0.7921 & 0.6939 & 0.7276 & 0.7287 & 0.7473 & 0.4926 & 0.5171 & 0.5254 & 0.5298 \cr	
  & SRLCH~\cite{DASSF2018}  & \textbf{0.7132} & \textbf{0.7184} & 0.7330 & 0.7437 & 0.6004 & 0.5796 & 0.6342 & 0.6053 & 0.5175 & 0.5346 & 0.5423 & 0.5470 &  0.3467	& 0.3466 & 0.3469 & 0.3471  \cr
   \cline{2-18}
		 & MTFH\_rnd & {0.7037} & 0.7150 & \textbf{0.7365} & \textbf{0.7399} & 0.7965 & 0.8067 & \textbf{ 0.8198} & 0.8303 & 0.7486 & 0.7760 & 0.7912 & {0.7938} & 0.6788 & 0.6980 & 0.7213 & 0.7201\cr
		 & MTFH\_km & 0.7020 & 0.7134& 0.7339 & 0.7368 & \textbf{0.8044} & \textbf{0.8146} & 0.8172 & \textbf{0.8352} & \textbf{0.7567} & \textbf{0.7797} & \textbf{0.7945} & \textbf{0.8044} & \textbf{0.6973}  & \textbf{0.7096} & \textbf{0.7326} & \textbf{0.7307} \cr
\hline
\end{tabular}
\end{center}
\label{tab:map}
\end{table*}
\begin{figure*}[t]
\begin{center}
   \includegraphics[width = 18.3cm]{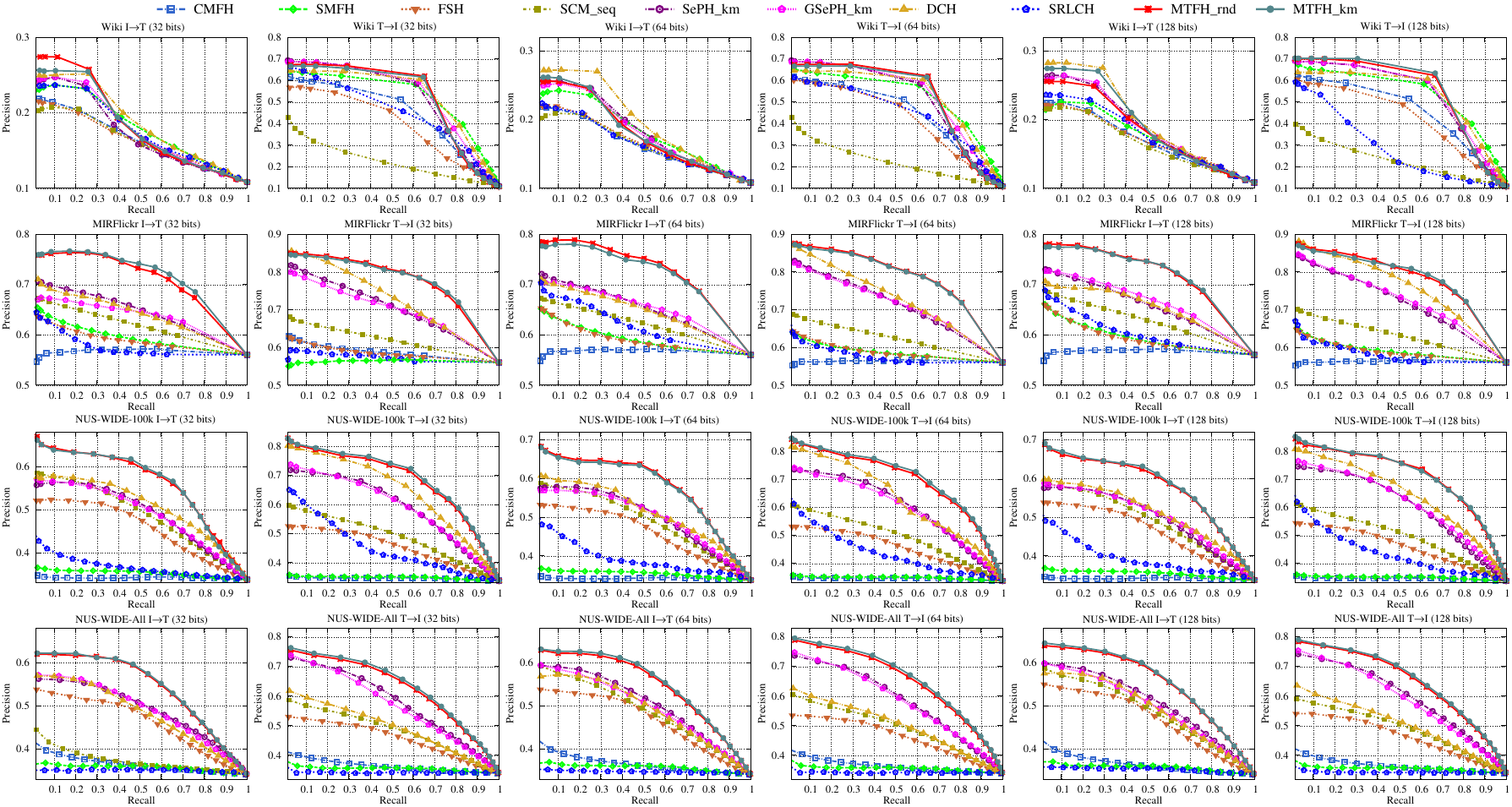}
\end{center}
\vspace{-0.4cm}
   \caption{Precision-recall curves obtained by different approaches and  tested on different datasets, in which the representative code lengths, \emph{i.e.}, 32, 64 and 128 bits, are selected for evaluation.}
\label{fig:PRcurve}
\end{figure*}

\subsection{Baseline Methods}
As surveyed in Section~\ref{relatedworks}, there exist many cross-modal hashing works. It is noted that the recent deep cross-modal hashing methods integrate the high-level feature learning and hash learning together, and our framework is totally different from those works. In that sense, it is really difficult to perform a relatively fair and meaningful comparison with these approaches appropriately. Specifically, we compare the proposed MTFH with eight well known
cross-modal hashing methods, including two unsupervised methods, \emph{i.e.},  CMFH \cite{ding2016large} and FSH~\cite{xmu2017},  and six  supervised
approaches, \emph{i.e.}, SMFH~\cite{tang2016supervised}, SCM~\cite{zhang2014large}, SePH~\cite{lin2015semantics}, GSePH \cite{mandal2017generalized}, DCH~\cite{xu2017learning} and SRLCH~\cite{DASSF2018}.  Those algorithms have been briefly introduced in Section~\ref{relatedworks} and  considered to be the current
state-of-the-arts in cross-modal hash learning.  Note that, some other competitive works are already reported within these works.

For the selected baselines,  we utilize the source codes kindly provided by the respective authors. The parameters are initialized as the authors have given in their original papers. As SePH~\cite{lin2015semantics} and SMFH~\cite{tang2016supervised} are computationally expensive, it is difficult to learn their corresponding hash functions on a larger training set. For the implementation of these two works, we follow their data processing suggestions and  sample a subset of 5000 instances, respectively from the retrieval sets of larger MIRFlickr and NUS-WIDE datasets, to form the training sets. For the other baselines, the training samples are initialized as the same as in the data description.
All the experiments are implemented using MATLAB and conducted on a computer running at an Intel Xeon$\textregistered$ E5-2609 1.90GHz processer with 128 GB memory. In the experiments, we perform five runs for each algorithm and take the average performance for illustration.

\subsection{Results of Equal Hash Length Encoding}\label{equallength}

As surveyed in Section~\ref{relatedworks}, almost all existing cross-modal hashing methods choose either unified or equal-length hash codes for multi-modal data representation. For fair comparison, we first set $q_1{=}q_2$ to learn the equal-length hash codes and vary the hash length from 16 to 128 bits (\emph{i.e.}, 16, 32, 64 and 128). Meanwhile, we  select both random (\textbf{rnd}) and k-means (\textbf{km}) sampling scheme in kernel logistic regression, and record the mAP scores on all four benchmark datasets.
Table~\ref{tab:map} displays the quantitative comparisons  of cross-modal retrieval performances with state-of-the arts baselines, while Fig.~\ref{fig:PRcurve} shows their precision-recall curves. It can be found that the proposed MTFH approach has achieved the comparable cross-modal retrieval performances in different hash length settings, and outperformed  most  baselines, \emph{i.e.}, CMFH~\cite{ding2016large}, SMFH~\cite{tang2016supervised},  FSH~\cite{ding2016large}, SCM~\cite{zhang2014large}, SePH~\cite{lin2015semantics} and GSePH~\cite{mandal2017generalized}.

For the small Wiki dataset, DCH~\cite{xu2017learning} has yielded  very competitive  mAP scores in I$\rightarrow$T task (\emph{i.e.}, 32, 64 and 128 bits), while SRLCH~\cite{DASSF2018} has resulted the larger mAP scores in T$\rightarrow$I task (\emph{i.e.}, 16 and 32 bits). However, their retrieval performances often  degrade on the larger datasets. Comparatively speaking, the proposed MTFH approach has delivered very competitive cross-modal retrieval performance on the Wiki dataset, and simultaneously yielded the best retrieval performance on the larger datasets.  The main reason lies that the Wiki dataset is a single-label dataset, while the other datasets are multi-label databases. For single-label dataset, some examples belonging to only one semantic label may have significantly different features. Under such circumstances, the features can be utilized to increase the discrimination power of hash code learning. Therefore, DCH and SRLCH are designed to jointly learn the hash functions and unified binary codes, which can produce very promising results on the Wiki dataset. For the multi-label dataset, the semantic labels are able to depict each instance, and the modality-specific hash codes derived from the proposed MTFH approach are more semantically meaningful than those generated from DCH and SRLCH.  As a result, the proposed MTFH has yielded the best retrieval performance on the larger datasets. For T$\rightarrow$I task, the mAP scores obtained by the proposed MTFH\_km approach  are higher than 0.80 and 0.75, respectively evaluated on the MIRFlickr and NUS-WIDE-100k datasets. For the largest  NUSWIDE-All dataset,  the hash codes of out-of-sample data can be well obtained  and the proposed MTFH method has also delivered the best cross-modal retrieval performances. The main superiorities contributed to these very competitive performances are three-fold: 1)  The modality-specific hash codes derived from MTFH are more discriminative and interpretable to characterize the heterogeneous data samples, while the unified hash representation may degrade their representation capability to represent both modalities.  2) MTF is more beneficial for revealing the latent structures within the heterogeneous samples, which can well characterize the native relations between data samples within the same modality and correlate the semantics between heterogeneous samples. Accordingly, the hash codes learned by the MTFH are more semantically meaningful than that generated by traditional matrix bi-factorization methods~\cite{ding2016large,mandal2017generalized}. 3) The hash functions learned from the discriminative hash codes are more efficient for mapping from features to hash codes, whereby the hash codes for out-of-sample data can be well computed.

\begin{figure*}[t]
\begin{center}
   \includegraphics[width = 18.3cm]{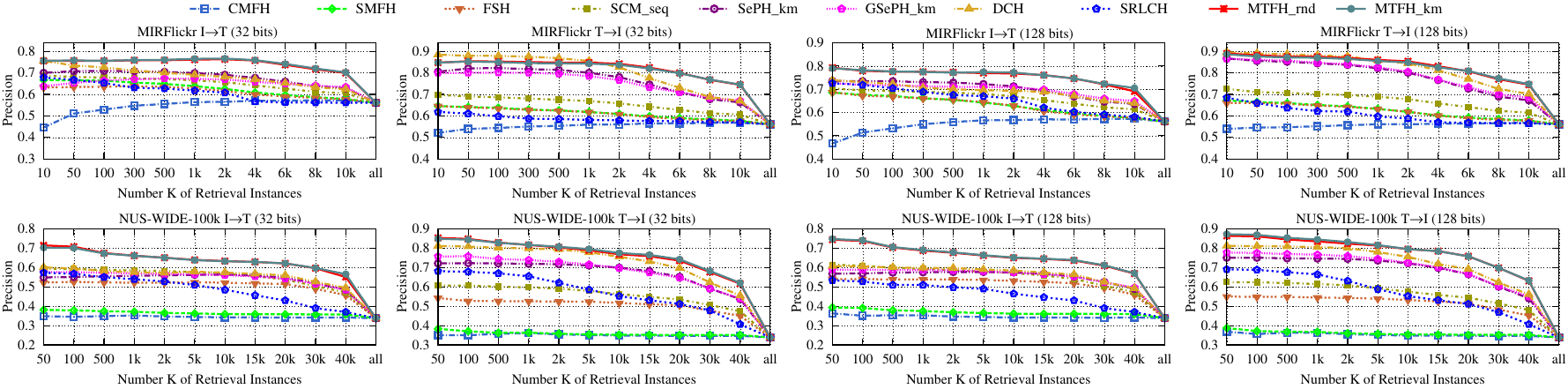}
\end{center}
\vspace{-0.4cm}
   \caption{The representative topK-precision curves tested on MIRFlickr and  NUS-WIDE-100k datasets.}
\label{fig:topK}
\end{figure*}


As suggested in~\cite{xu2017learning}, we further utilize  mAP@K and topK-precision  to measure the retrieval performances within the top-ranked $\text{K}$ retrieved items. Specifically, topK-precision reflects the change of precision with respect to the number of top-ranked K instances presented to the users.
For these two metrics, larger value generally indicates the better retrieval performance. As displayed in Table~\ref{tab:map50}, we record the representative mAP@50 values in typical MIRFlickr and NUS-WIDE-100k datasets. It can be found that the proposed MTFH approach yields the comparable  mAP@50 values with DCH when tested on  MIRFlickr, and outperforms the state-of-the-art baselines on NUS-WIDE-100k.   Meanwhile, the representative  topK-precision curves (\emph{i.e.}, 32 and 128 bits) are shown in Fig.~\ref{fig:topK}, it can be seen that the proposed MTFH  method always yields the highest precision  scores than the baselines with the number of retrieved instances (K) changes. This indicates that the proposed  MTFH approach is capable of returning much more similar samples at the beginning, which is very important for a practical retrieval system. Therefore, the proposed MTFH associated with equal hash length setting is very competitive to the state-of-the-art cross-modal retrieval baselines.

\begin{table}[t]
\caption{Representative cross-modal retrieval performance (mAP@50) obtained by  different approaches, and the best results are highlighted in bold.}\setlength{\tabcolsep}{0cm}
\vspace{-0.5cm}
\begin{center}
\begin{tabular}{|p{1.9cm}<{\centering}|p{0.88cm}<{\centering}p{0.88cm}<{\centering}|p{0.88cm}<{\centering}p{0.88cm}<{\centering}|p{0.88cm}<{\centering}p{0.88cm}<{\centering}|p{0.88cm}<{\centering}p{0.88cm}<{\centering}|}
\hline
\multirow{3}{*}{Method} &    \multicolumn{4}{c|}{MIRFlickr} &  \multicolumn{4}{c|}{NUS-WIDE-100k}    \\
 \cline{2-9}
&     \multicolumn{ 2}{c|}{I$\rightarrow$T} &    \multicolumn{ 2}{c|}{T$\rightarrow$I}   &    \multicolumn{ 2}{c|}{I$\rightarrow$T} & \multicolumn{ 2}{c|}{T$\rightarrow$I}    \\
 \cline{2-9}
 & {32} & {128} & {32} & {128} & {32} & {128} & {32} & {128} \\
\hline
{CMFH~\cite{ding2016large} } &    0.5257  &    0.5798  &    0.5701  &    0.5846  &    0.4026  &    0.4200  &    0.4052  &    0.4267  \\
{SMFH~\cite{tang2016supervised}} &    0.6915  &    0.7052  &    0.6691  &    0.6928  &    0.4291  &    0.4327  &    0.4025  &    0.4240  \\
{FSH~\cite{xmu2017} } &    0.6804  &    0.6960  &    0.6744  &    0.6951  &    0.5734  &    0.5706  &    0.6024  &    0.5883  \\
{SCM\_orth~\cite{zhang2014large}} &    0.6510  &    0.6593  &    0.6682  &    0.6394  &  0.5168  &    0.4540  &    0.5124  &    0.4594  \\
{SCM\_seq~\cite{zhang2014large}} &    0.7061  &    0.7217  &    0.7160  &    0.7395  &    0.6230  &    0.6464  &    0.6366  &    0.6509  \\
{SePH\_rnd~\cite{lin2015semantics}} &    0.7260  &    0.8546  &    0.8301  &    0.8652  &    0.5813  &    0.5993  &    0.7299  &    0.7635  \\
{SePH\_km~\cite{lin2015semantics}} &    0.7237  &    0.8563  &    0.8276  &    0.8703  &    0.5798  &    0.5956  &    0.7335  &    0.7681  \\
{GSePH\_rnd~\cite{mandal2017generalized}} &    0.6773  &    0.8370  &    0.8106  &    0.8655  &    0.5996  &    0.6133  &    0.7702  &    0.7873  \\
{GSePH\_km~\cite{mandal2017generalized}} &    0.6679  &    0.8398  &    0.8119  &    0.8727  &    0.6082  &    0.6166  &    0.7808  &    0.7936  \\
{DCH~\cite{xu2017learning}} &    0.7723  &    0.8885  & { \bf 0.8923 } & { \bf 0.9013 } &    0.6396  &    0.6301  &    0.8231  &    0.8171  \\
{SRLCH~\cite{DASSF2018}} &    0.7480  &    0.7849  &  0.8284  & 0.7675 &    0.7670  &    0.8645  &    0.7510  &    0.8495  \\
\hline
{TMFH\_rnd} &    0.7713  & {\bf 0.8932 } &    0.8619  &    0.8992  & { \bf 0.7337 } &    0.7723  & { \bf 0.8687 } &    0.8766  \\
{ TMFH\_km} & { \bf 0.7739 } &    0.8887  &    0.8624  &    0.8935  &    0.7272  & { \bf 0.7753 } &    0.8616  & { \bf 0.8814 } \\
\hline
\end{tabular}
\end{center}
\label{tab:map50}
\end{table}
\begin{figure*}[t]
\centering
   \includegraphics[width=18.3cm]{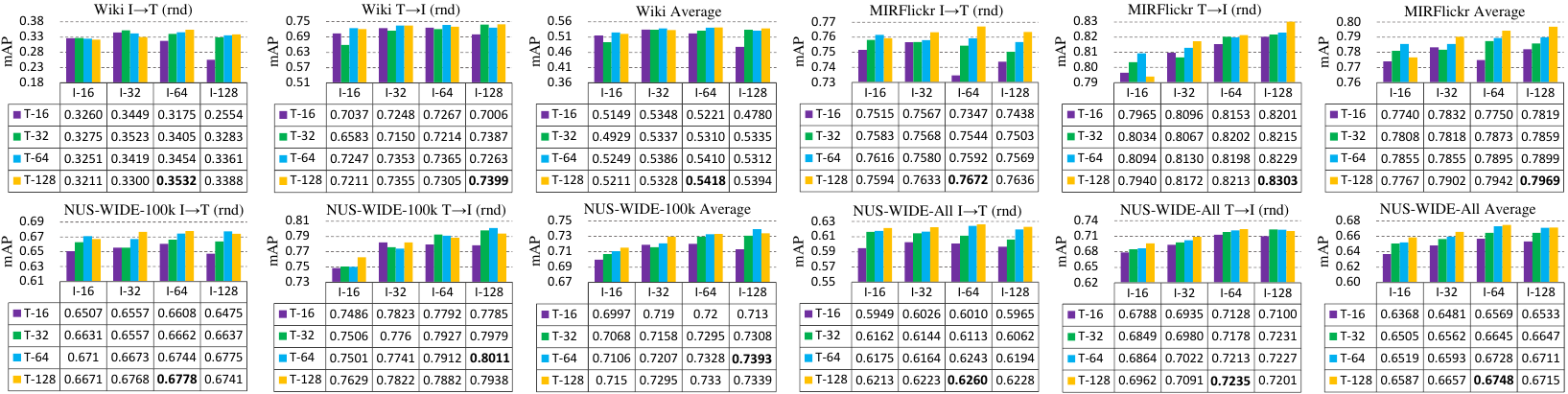}\\
   \vspace{-0.2cm}
   \caption{Cross-modal retrieval results obtained by the proposed MTFH with varying hash length settings, and the best results are highlighted in bold.}
\label{fig:diff}
\end{figure*}

\subsection{Results of Unequal Hash Length Encoding}\label{varinglength}

\begin{figure}[t]
\centering
   \includegraphics[width=9cm]{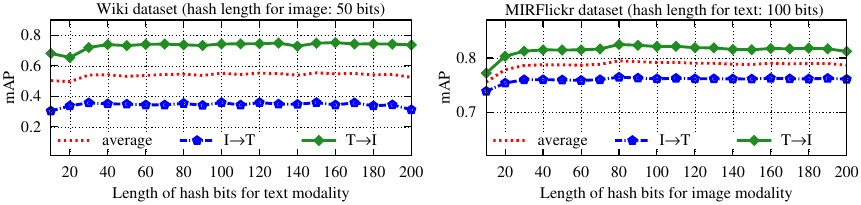}
   \vspace{-0.6cm}
   \caption{Cross-modal retrieval results by fixing the hash length of one modality and varying the hash length of another modality.}
\label{fig:diff2}
\end{figure}

The proposed MTFH framework is the first attempt to generate varying hash codes of different lengths for multi-modal data representation.
 To  validate the flexibility and effectiveness of the proposed framework, we set $q_1{\neq}q_2$ and conduct a series of experiments with unequal hashing length settings, \emph{e.g.}, the  hash lengths corresponding to image and text modalities are set at 16 (I-16) and 32 (T-32) bits, respectively. The mAP values obtained by unequal hash length settings are displayed in Fig. \ref{fig:diff}, it can be seen that  the best retrieval performances are not always achieved by the equal hash length representations, and varying hash length encoding scheme has also delivered very competitive cross-modal retrieval performance. For instance, if the MTFH\_rnd method is selected, the best I$\rightarrow$T retrieval results tested on the MIRFlickr and NUS-WIDE-100k datasets are generated by hash pair I-64\&T-128. The similar results can be also found in  their average retrieval performances. The main reason lies that the feature dimensions corresponding to the image and text modalities are different, and such difference makes the varying hash length encoding scheme to be efficient for heterogeneous data representation.  Further, we record the mAP scores by fixing the hash length of one modality to be constant and varying the hash bits of another modalities to be different. Typical examples are shown in Fig.~\ref{fig:diff2}, it can be found that  the larger code length does not always improve the cross-modal retrieval performance and the optimum retrieval results are not usually achieved by the equal hash length encoding scenarios. It is noted that the varying hash length encoding of different modalities has delivered the comparable and even better retrieval performances. For instance, the hash pair I-80\&T-100 has achieved the better retrieval performances (\emph{i.e.}, larger mAP scores) than that obtained by hash pair I-100\&T-100, when tested on MIRFlickr dataset. That is, the proposed MTFH method can shorten the hash bits of one modality to index relevant samples without degrading the performances. Therefore, the hash representations of heterogeneous modalities encoded by different code lengths are feasible and meaningful, especially when the feature dimensions of heterogeneous modalities differ sharply.


Further, we evaluate the recall rates by using unequal hash lengths. As the feature dimension of text modality in the Wiki dataset is only equal to 10, we fix the hash length of image modality to be 128, and report the recall rates by varying the hash bits of text modality from 16 to 128. Meanwhile, we also record the recall scores with equal hash length encoding scenarios, \emph{i.e.}, I-16\&T-16, I-32\&T-32 and I-64\&T-64. As shown in Table~\ref{tab:recall}, it can be found that the best recall rates are not achieved by the equal hash length representations. For instance, the hash pair I-128\&T-64 has achieved the best recall rate of  I$\rightarrow$T task when the top 500 instances are searched. The main reason lies in that the image-text pairs are not always optimally encoded by the equal hash lengths due to their different sample size and distinct feature dimensions, thereby the strictly equalized hash length setting cannot guarantee the learned binary codes to be semantically discriminative for heterogeneous data representation. Another possible reason is that a bit long hash representation of  low-dimensional
text data may result in low recall, since the collision probability that two codes fall into the same hash
bucket may decrease exponentially as the code length increases. It is noted that  the recall rates are not improved when we search the relevant samples with higher number of bits, \emph{e.g.}, I-128\&T-128. Under the circumstances, the proposed MTFH method incorporating with less hash bits could save the storage memory, which we will discuss it in Section~\ref{extension}. Therefore, the proposed varying hash length encoding scheme is beneficial to produce more effective hash code for
 heterogeneous data representation and performance improvements.  More importantly,  the proposed cross-modal retrieval framework is particularly adaptive to an even more challenging scenario, \emph{i.e.}, the hash representations from heterogeneous modalities are encoded and stored by different lengths in the database. The experimental results have shown its flexibility with outstanding performances.


\begin{table}[!t]
\caption{Recall rates obtained by MTFH and tested with different hash lengths on Wiki dataset. The best results are highlighted in bold.}\setlength{\tabcolsep}{0.05cm}
\vspace{-0.5cm}
\begin{center}
\begin{tabular}{|c|c|c|c|c|c|c|c|c|}
\hline
\multicolumn{1}{|c|}{\multirow{2}*{ Bit length}} & \multicolumn{8}{c|}{Recall rates of I$\rightarrow$T task with different ranking instances.} \\
\cline{2-9}
\multicolumn{1}{|c|}{} &  { 50} &  {100} &  {250} &  {500} &  {750} & {1000} & {1500} & {2000} \\
\hline
  {I-16\&T-16} &    0.0489  &    0.0981  &    0.2370  &    0.3781  &    0.4891  &    0.5952  &    0.7802  &    0.9440  \\
\hline
  {I-32\&T-32} &    0.0507  &    0.1021  &    0.2401  &    0.3828  &    0.5109  &    0.6199  &    0.8134  &    0.9536  \\
  \hline
  {I-64\&T-64} &    0.0530  &    0.1058  &    0.2499  &    0.3959  &    0.5185  &    0.6302  &    0.8116  &    0.9587  \\
  \hline
 {I-128\&T-128} &    0.0542  &    0.1076  &    0.2515  &    0.3876  &    0.5131  &    0.6288  &    0.8121  &    0.9554  \\
\hline
  {I-128\&T-16} &    0.0366  &    0.0741  &    0.1807  &    0.3041  &    0.4117  &    0.5272  &    0.7423  &    0.9444  \\
\hline
  {I-128\&T-32} &    0.0514  &    0.1028  &    0.2417  &    0.3935  &    \textbf{0.5241}  &    \textbf{0.6340}  &    \textbf{0.8327}  &    \textbf{0.9640}  \\
\hline
  {I-128\&T-64} &    \textbf{0.0565} &   \textbf{0.1133}  &   \textbf{0.2669}  &    \textbf{0.4093}  &    0.5240  &    0.6263  &    0.8007  &    0.9503  \\
\hline
\end{tabular}
\end{center}
\label{tab:recall}
\end{table}


\subsection{Results of the Unpaired Scenario}

The experiments reported in Section~\ref{equallength} and \ref{varinglength} mainly focus on the paired multi-modal data collections. For the unpaired data collections, we further evaluate the proposed MTFH method on both single-label unpaired (SL-U) and multi-label unpaired (ML-U)  scenarios. That is, multi-modal data from different modalities may not have one-to-one correspondence, \emph{e.g.}, 100 images and 90 text documents share the same semantic tag ``flower".

\begin{table}[!t]
\caption{Retrieval results (mAP) of unpaired multi-modal data collections, and the best results are highlighted in bold.}\setlength{\tabcolsep}{0.14cm}
\vspace{-0.6cm}
\begin{center}
\begin{tabular}{|c|c| c|c|c|c|}
\hline
	\multicolumn{2}{|c|}{\multirow{2}*{Method}} & \multicolumn{2}{c|}{Wiki (I$\rightarrow$T/T$\rightarrow$I)} & \multicolumn{2}{c|}{MIRFlickr (I$\rightarrow$T/T$\rightarrow$I)} \cr \cline{3-6}
	\multicolumn{2}{|c|}{} & unpair-1 & unpair-2 & unpair-1 & unpair-2 \cr \cline{3-6}
	\hline
	\multicolumn{2}{|c|}{CCA~\cite{hardoon2004canonical}} & 0.176/0.156 & 0.178/0.154 & 0.581/0.579 & 0.581/0.579 \\
	\hline
	\multicolumn{2}{|c|}{IMH \cite{song2013inter}} & 0.176/0.156 & 0.178/0.154 & 0.581/0.579 & 0.581/0.579 \\
	\hline
	\multirow{4}{*}{CMFH~\cite{ding2016large}}
		& 16 & 0.196/0.496 & 0.205/0.452 & 0.567/0.564 & 0.567/0.563 \\
		& 32 & 0.204/0.509 & 0.231/0.491 & 0.568/0.566 & 0.568/0.564 \\
		& 64 & 0.215/0.532 & 0.232/0.492 & 0.568/0.565 & 0.568/0.564 \\
		& 128 & 0.220/0.534 & 0.240/0.507 & 0.568/0.566 & 0.568/0.564 \\
	\hline
	\multirow{4}{*}{GSePH~\cite{mandal2017generalized}}
		& 16 & 0.257/0.453 & 0.268/0.422 & 0.651/0.631 & 0.653/0.645 \\
		& 32 & 0.273/0.477 & 0.279/0.438 & 0.648/0.633 & 0.658/0.635 \\
		& 64 & 0.283/0.483 & 0.298/0.456 & 0.665/0.665 & 0.675/0.663 \\
		& 128 & 0.288/0.490 & 0.292/0.466 & 0.676/0.670 & 0.681/0.668 \\
	\hline
	\multirow{4}{*}{DCH~\cite{xu2017learning}}
		& 16 & 0.324/0.692 & 0.304/0.636 & 0.661/0.745 & 0.675/0.741 \\
		& 32 & 0.336/0.717 & 0.354/0.668 & 0.657/0.738 & 0.673/0.737 \\
		& 64 & 0.349/0.716 & 0.379/0.683 & 0.666/0.760 & 0.679/0.750 \\
		& 128 & 0.347/0.723 & \textbf{0.384}/0.690 & 0.686/0.796 & 0.690/0.771 \\
	\hline
	\multirow{4}{*}{MTFH}
		& 16 & 0.329/0.711 & 0.316/0.727 & 0.733/0.759 & 0.754/0.808 \\
		& 32 & 0.342/0.727 & 0.343/0.736 & 0.757/0.811 & 0.757/0.819 \\
		& 64 & \textbf{0.355}/\textbf{0.734} & 0.330/\textbf{0.749} & 0.761/0.820 & 0.759/\textbf{0.827} \\
		& 128 & 0.340/0.707 & {0.365}/0.742 & \textbf{0.765}/\textbf{0.832} & \textbf{0.767}/0.824 \\
		\hline
\end{tabular}
\end{center}
\label{tab:unpair}
\end{table}

For SL-U,  each data point is associated with a single label, but there does not exist one-to-one correspondence between the data of two modalities. In this case,  the Wiki dataset is selected for evaluation. Similar to~\cite{mandal2017generalized}, we keep the text modality unchanged and randomly select 90\% of images as `unpair-1' and vice verse as `unpair-2'. For ML-U, each data point is associated with multiple labels, but there also does not exist one-to-one correspondence between the data of two modalities. In this case, MIRFlickr dataset is selected for evaluation, and we follow the same organizing way as SL-U to form the unpaired data from  MIRFlickr dataset. Specifically, the training set itself serves as the retrieval set while the query set is kept unchanged as in the paired cases.  Except for GSePH \cite{mandal2017generalized},  other cross-modal retrieval algorithms developed for paired multi-modal collections are not applicable to handle this unpaired scenario. We  follow the data processing ways in \cite{mandal2017generalized} to artificially construct the paired training sets and heuristically implement the CCA \cite{hardoon2004canonical}, IMH \cite{song2013inter}, CMFH~\cite{ding2016large} and DCH~\cite{xu2017learning} for meaningful comparison. In GSePH and MTFH, the random (\textbf{rnd}) sampling scheme is selected in kernel logistic regression.

The cross-modal retrieval performances tested on unpaired data are shown in Table \ref{tab:unpair}. It can be observed that  CCA and IMH methods have delivered relatively lower mAP scores, while  CMFH and GSePH approaches have also  degraded their retrieval performances in unpaired multi-modal data collections.  By contrast, our proposed MTFH method significantly outperforms these baseline methods.  For I$\rightarrow$T task, the mAP values obtained by GSePH  and tested on MIRFlickr dataset  drop slightly on both unpaired tasks, which are all less than 0.69.  Relatively speaking, our proposed MTFH method yields the very competitive I$\rightarrow$T performances and the corresponding mAP values are higher than 0.73.  By artificially pairing the training samples, we notice that DCH has achieved the promising retrieval performances, especially for the T$\rightarrow$I task on the Wiki dataset. However, the mAP scores obtained by DCH were relatively unstable when tested on MIRFlickr dataset. In contrast to this, our proposed MTFH has achieved very stable performance on MIRFlickr dataset and the corresponding mAP values are always higher than the results obtained by DCH. That is, our proposed MTFH approach can not only handle various unpaired multi-modal data collections, but also produce relatively stable retrieval performance on different retrieval tasks.


%

\subsection{Results of Single-modal Retrieval}
\begin{figure}[!t]
\begin{center}
   \includegraphics[width=9cm]{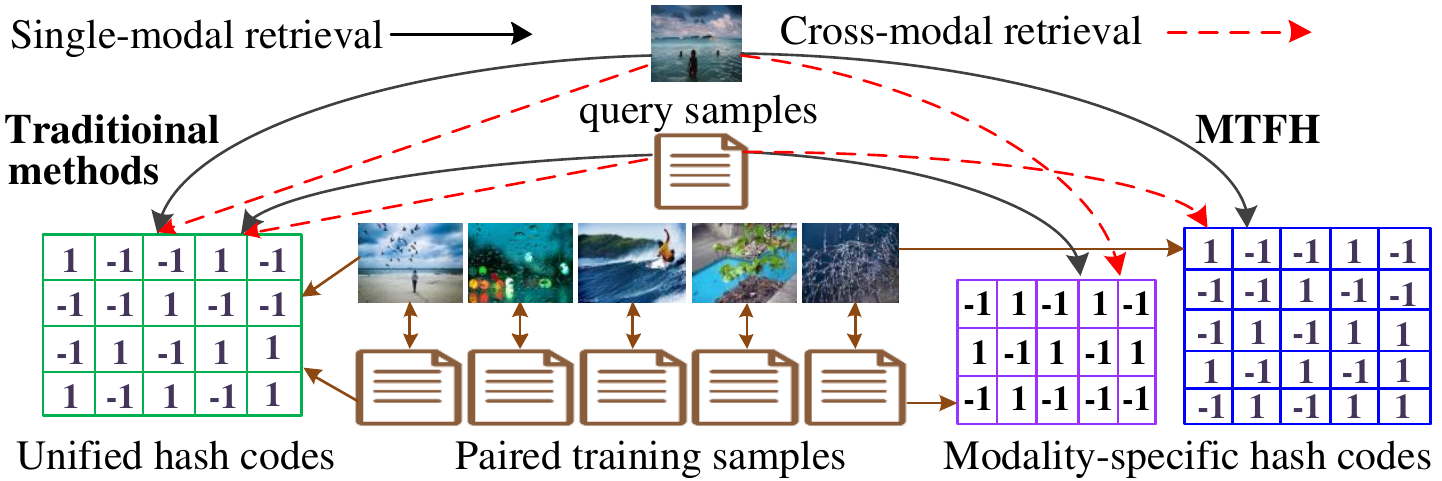}
\end{center}
\vspace{-0.4cm}
   \caption{The illustration of different hash  representation for single-modal retrieval and cross-modal retrieval tasks.}
\label{fig:single}
\end{figure}

The majority of existing cross-modal hashing methods often learn unified hash codes to characterize the paired multi-modal data. As shown in Fig.~\ref{fig:single}, if the unified hash codes are utilized to represent the heterogeneous data points,  these approaches naturally yield the same retrieval performance in both single-modal and cross-modal retrieval tasks. In contrast to this, the hash codes of heterogeneous modalities derived from the proposed MTFH approach are different, and these learned modality-specific hash codes can be well utilized for single-modal retrieval. As indicated in FSH~\cite{xmu2017}, the integration of multiple modalities often improves the search performance, and we further  evaluate our learned hash codes on single-modal retrieval, \emph{i.e.}, image-to-image (I$\rightarrow$I) and text-to-text (T$\rightarrow$T). Specifically, the random (\textbf{rnd}) sampling scheme is adopted in kernel logistic regression. Meanwhile, we select three competing single-modal hashing baselines, \emph{i.e.}, Iterative Quantization (ITQ)~\cite{singleAdd1}, Scalable Graph Hashing (SGH)~\cite{singleAdd2} and Fast Supervised Discrete Hashing (FSDH)~\cite{fasthash2018}, and one representative cross-modal hashing (non-unified hash representation), \emph{i.e.}, FSH~\cite{xmu2017}, for meaningful comparison.   Note that, the other unified hash representations are not selected because these works naturally yield the same retrieval performances in both single-modal retrieval and cross-modal retrieval, as shown in Table~\ref{tab:map}.

\begin{table}[!t]
\caption{Results (mAP) of single-modal retrieval on paired multi-modal data, and the best results are highlighted in bold.}\setlength{\tabcolsep}{0.12cm}
\vspace{-0.5cm}
\begin{center}
\begin{tabular}{|c|c|c|c|c|}
\hline
	\multicolumn{1}{|c|}{\multirow{2}*{Method}}  & \multicolumn{1}{c|}{\multirow{2}*{Bit length}} & \multicolumn{1}{c|}{Wiki } & \multicolumn{1}{c|}{MIRFlickr} & \multicolumn{1}{c|}{NUS-WIDE-100k} \cr \cline{3-5}
\multicolumn{1}{|c|}{} &\multicolumn{1}{c|}{} & I$\rightarrow$I/T$\rightarrow$T & I$\rightarrow$I/T$\rightarrow$T & I$\rightarrow$I/T$\rightarrow$T \cr \cline{3-5}
	\hline
\multirow{3}{*}{ITQ~\cite{singleAdd1}}
		& 32 & 0.114/0.414 & 0.573/0.583 & 0.381/0.353 \\
        & 64 & 0.113/0.414 & 0.552/0.578 & 0.381/0.349  \\
	    & 128 & 0.111/0.414 & 0.575/0.562 & 0.383/0.349  \\
		\hline
\multirow{3}{*}{SGH~\cite{singleAdd2}}
        & 32 & 0.121/0.440 & 0.582/0.579 & 0.338/0.373 \\
        & 64 & 0.120/0.460 & 0.583/0.581 & 0.339/0.371  \\
	    & 128 & 0.120/0.486 & 0.583/0.579 & 0.339/0.369  \\
		\hline
\multirow{3}{*}{FSDH~\cite{fasthash2018}}
        & 32 & 0.215/0.555& 0.663/0.694 & 0.492/0.523 \\
        & 64 & 0.245/0.610 & 0.661/0.699 & 0.483/0.518  \\
	    & 128 & 0.276/0.667 & 0.672/0.715 & 0.511/0.556  \\
		\hline
\multirow{3}{*}{FSH~\cite{xmu2017}}
        & I-32\&T-32 & 0.161/0.519 & 0.592/0.605 & 0.462/0.521 \\
        & I-64\&T-64 & 0.165/0.520 & 0.590/0.604 & 0.469/0.538  \\
	    & I-128\&T-128 & 0.167/0.536 & 0.593/0.607 & 0.467/0.537  \\
		\hline
\multirow{9}{*}{\makecell[cc]{MTFH}}
	    &I-32\&T-32 & 0.363/0.738 & 0.748/0.823 & 0.662/0.797 \\
        & I-64\&T-64 & 0.363/0.748 & 0.760/0.820 & 0.675/0.805 \\
	    & I-128\&T-128 & 0.373/0.740 & \textbf{0.768}/0.830 & 0.683/0.795  \\
		\cline{2-5}
	    & I-32\&T-64 & 0.355/0.739 & 0.754/0.819 & 0.666/0.793 \\
        & I-32\&T-128 & 0.366/0.736 & 0.759/0.827 & 0.665/\textbf{0.809}  \\
	    & I-64\&T-32 & 0.362/0.744 & 0.759/0.816 & 0.673/0.780  \\
	    & I-64\&T-128 & \textbf{0.383}/0.746 & 0.761/\textbf{0.832} & 0.678/0.795 \\
        & I-128\&T-32 & 0.378/0.734 & 0.763/0.811 & 0.679/0.782  \\
	    & I-128\&T-64 & 0.376/\textbf{0.749} & 0.763/0.823 & \textbf{0.690}/0.796  \\
		\hline
\end{tabular}
\end{center}
\label{tab:singleC}
\end{table}

Table~\ref{tab:singleC} shows the single-modal retrieval results on representative datasets. It can be observed that hash codes of equal lengths derived from the proposed MTFH method have always delivered a better single-modal retrieval performance than that generated from both representative single-modal hashing methods (\emph{i.e.}, ITQ~\cite{singleAdd1}, SGH~\cite{singleAdd2} and FSDH~\cite{fasthash2018}) and  non-unified hash representation method (\emph{i.e.}, FSH~\cite{xmu2017}). Meanwhile, as compared in Table~\ref{tab:map}, the single-modal retrieval performances obtained by MTFH are generally better than most results that produced by unified hash representations (\emph{e.g.}, CMFH~\cite{ding2016large}, SePH~\cite{lin2015semantics} and GSePH \cite{mandal2017generalized}). This demonstrates that the proposed MTFH framework is able to produce more distinguished binary codes for both heterogeneous modalities, which subsequently improves the single-modal retrieval performance. That is, the proposed MTFH method not only exhibits the flexibility in cross-modal retrieval, but also shows very competitive performance in single-modal retrieval task.


Further, the proposed MTFH framework is able to jointly learn the modality-specific hash codes with different hash length settings, and some derived hash codes with varying lengths have also boosted the single-modal retrieval performance. For instance, the learned multi-modal hash codes, \emph{e.g.}, I-128\&T-64, yield the best I$\rightarrow$I  retrieval performance on the NUS-WIDE-100k dataset. That is,  the hash codes derived from the couple lengths, \emph{i.e.}, I-128\&T-64, are more semantically meaningful for  single-modal retrieval on NUS-WIDE-100k dataset. The experimental results have shown its scalability in single-modal retrieval tasks.

\subsection{Results of CNN Visual Features}
With the development of convolutional neural network (CNN),  the visual features obtained from the pretrained or fine-tuned CNN models have demonstrated to be effective for cross-modal retrieval~\cite{wei2017cross}, and the improved performance can be achieved  based on  classic cross-modal retrieval methods, such as CCA~\cite{hardoon2004canonical} and three-view CCA~\cite{gong2014multi}.  Accordingly, we evaluate the proposed MTFH on the Wiki, Pascal Sentence~\cite{rashtchian2010collecting} and  Pascal VOC 2007~\cite{everingham2010pascal} datasets, and their CNN visual features are publicly shared by work~\cite{wei2017cross}. Specifically, the off-the-shelf fine-tuned CNN visual features, \emph{i.e.}, FT-fc7, are selected for evaluation~\cite{wei2017cross}. Meanwhile, we carefully implement  CCA~\cite{hardoon2004canonical}, three view CCA (T-V CCA)~\cite{gong2014multi}, deep Semantic Matching (deep-SM)~\cite{wei2017cross}, CMFH~\cite{ding2016large}, SePH~\cite{lin2015semantics}, GSePH~\cite{mandal2017generalized} and DCH~\cite{xu2017learning} for comparison. Comparing with the hand-crafted visual features,  the dimensionality of CNN feature is large, \emph{i.e.}, 4096. Therefore, we typically set the code length to 32 and 128, and  equalize the hash length of two heterogeneous modalities for fair evaluation.

\begin{table}[!t]
\caption{Results (mAP) of cross-modal retrieval on CNN visual features, and the best results are highlighted in bold.}\setlength{\tabcolsep}{0.16cm}
\vspace{-0.3cm}
\begin{center}
\begin{tabular}{|c|c|c|c|c|}
\hline
	\multicolumn{2}{|c|}{\multirow{2}*{Method}} & \multicolumn{1}{c|}{Wiki } & \multicolumn{1}{c|}{Pascal Sentence } & \multicolumn{1}{c|}{Pascal VOC 2007} \cr \cline{3-5}
	\multicolumn{2}{|c|}{} & I$\rightarrow$T/T$\rightarrow$I & I$\rightarrow$T/T$\rightarrow$I & I$\rightarrow$T/T$\rightarrow$I \cr \cline{3-5}
	\hline
	\multicolumn{2}{|c|}{CCA~\cite{hardoon2004canonical}} & 0.272/0.287 & 0.307/0.372 & 0.635/0.643 \\
	\hline
	\multicolumn{2}{|c|}{T-V CCA~\cite{gong2014multi}} & 0.311/0.316 & 0.338/0.438 & 0.689/0.714  \\
	\hline
	\multicolumn{2}{|c|}{Deep-SM~\cite{wei2017cross}} & 0.398/0.354 & 0.446/0.478 & \textbf{0.823}/0.776  \\
	\hline
\multirow{2}{*}{CMFH~\cite{ding2016large}}
		& 32 & 0.184/0.265 & 0.323/0.424 & 0.382/0.703 \\
	    & 128 & 0.187/0.325 & 0.361/0.490 & 0.279/0.339  \\
		\hline
\multirow{2}{*}{SePH~\cite{lin2015semantics}}
  &32 &     0.476/0.734  &   0.497/0.690  & 0.749/0.877  \\
  &128 &     0.520/0.774  &  0.543/0.729  & 0.784/0.912\\
		\hline
\multirow{2}{*}{GSePH~\cite{mandal2017generalized}}
&32 &     0.494/0.762  &  0.428/0.574 &     0.763/0.900\\
&128 &     0.508/0.777  & 0.463/0.646 &     0.802/0.946\\
		\hline
\multirow{2}{*}{DCH~\cite{xu2017learning}}
		& 32 & 0.433/0.782 & 0.587/0.799 & 0.536/0.838 \\
	    & 128 & 0.456/0.793 & \textbf{0.605}/\textbf{0.801} & 0.577/0.876  \\
		\hline
	\multirow{2}{*}{MTFH}
		& 32 & \textbf{0.544}/0.724 & 0.594/0.779 & 0.749/0.883 \\
	    & 128 & 0.523/\textbf{0.809} & 0.604/0.787 & 0.805/\textbf{0.961}  \\
		\hline
\end{tabular}
\end{center}
\label{tab:CNNfeature}
\end{table}

The representative cross-modal retrieval performances evaluated on the fine-tuned CNN visual features are displayed in Table~\ref{tab:CNNfeature}, it can be observed that both of DCH~\cite{xu2017learning} and the proposed MTFH method yield the better retrieval performances than the results produced by other competing baselines, \emph{i.e.}, CCA~\cite{hardoon2004canonical}, T-V CCA~\cite{gong2014multi}, deep-SM~\cite{wei2017cross}, CMFH~\cite{ding2016large}, SePH~\cite{lin2015semantics} and GSePH~\cite{mandal2017generalized}. We notice that DCH~\cite{xu2017learning} has delivered very competitive mAP scores in Pascal sentence dataset (\emph{i.e.}, 128 bits), but its retrieval performance degrades on the Wiki and Pascal VOC 2007 datasets. Comparatively speaking, the proposed MTFH approach often boosts the retrieval performances in different hash length settings, and significantly outperforms most state-of-the-art baselines, especially on the Wiki and Pascal VOC 2007 datasets. For instance, the Wiki dataset is a very popular multi-modal dataset, and the CNN visual features can further benefit the cross-modal retrieval performance. If the hash length is set at 128 bits, the mAP scores obtained by MTFH are higher than 0.5  and 0.8, respectively, evaluated on I$\rightarrow$T and T$\rightarrow$I tasks. This demonstrates that  the learned hash projection functions can well map the CNN visual features into compact hash codes. That is, the proposed MTFH framework is applicable to various kinds of  sample features and the experimental results have demonstrated its efficiency.





\subsection{Effects of Discrete Optimization}
Within the proposed MTFH framework,  an efficient  discrete optimization
algorithm is proposed to jointly learn the modality-specific hash
codes without relaxation. Sine the relaxation scheme  may accumulate large quantization
error as the code length increases,  DCH~\cite{xu2017learning} utilizes a discrete
cyclic coordinate decent (DCC) algorithm to learn and update each hash bit in a cyclic order, which is evidently an approximate solution to the discrete hashing and may fall into a local minimum during the learning process. To alleviate this problem, we improve DCC and utilize the
E-RCD to derive the hash codes more reliably.

\begin{figure*}[!htp]
\begin{center}
   \includegraphics[width=18.2cm]{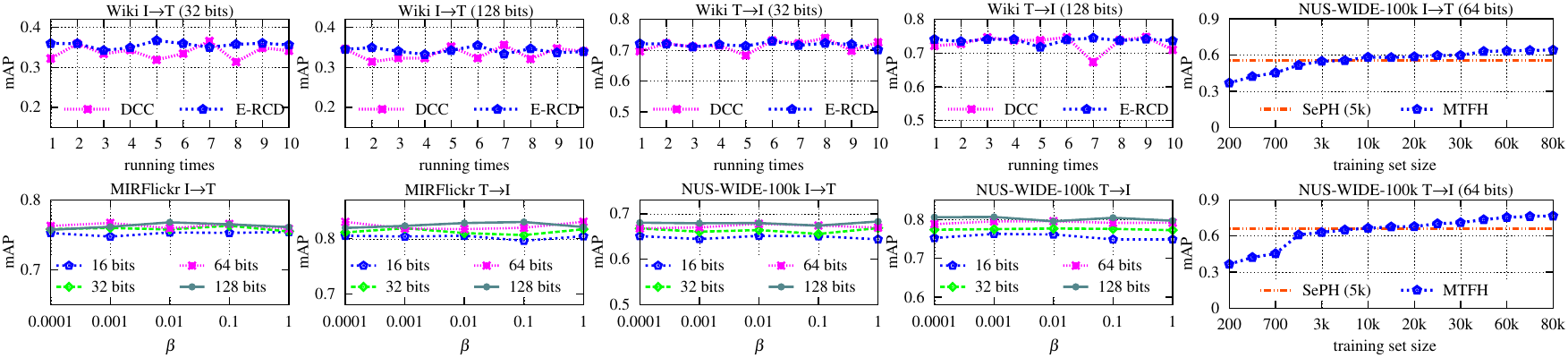}
\end{center}
\vspace{-0.4cm}
   \caption{Effects of different optimization schemes, parameter values and training set sizes. }
\label{fig:parameter}
\end{figure*}

Further, we compare DCC with the proposed E-RCD in solving the same objective
function, \emph{i.e.}, Eq.~(\ref{eq:final}). We take the paired Wiki dataset for testing, and learn the hash codes of equal lengths (\emph{i.e.}, 32 bits and 128 bits) for evaluation. As the solutions of both DCC and E-RCD
depend  on the initial values of model parameters, we  run ten times for both optimizations. Note that,  similar results can be also found in MIRFlickr and NUS-WIDE datasets, as well as other retrieval tasks (\emph{i.e.}, unequal hash length encoding,  unpaired multi-modal data collection, single-modal retrieval and CNN visual features). Fig.~\ref{fig:parameter} shows the changes of the corresponding mAP values tested by DCC
and E-RCD within ten trials, and Table~\ref{tab:optimization} displays their statistical properties. As compared in Table~\ref{tab:map}, the proposed MTFH framework solved by DCC directly also yields satisfactory performance in both retrieval tasks (I$\rightarrow$T and  T$\rightarrow$I), and always outperforms most state-of-the-art baselines, \emph{i.e.}, CMFH~\cite{ding2016large}, SMFH~\cite{tang2016supervised}, FSH~\cite{ding2016large}, SCM~\cite{zhang2014large}, SePH~\cite{lin2015semantics} and GSePH~\cite{mandal2017generalized}.  For instance, the average mAP values derived from 128 bits and computed from ten trials reach up to 0.3342 and 0.7284, respectively, evaluated on I$\rightarrow$T and  T$\rightarrow$I tasks.

As shown in Fig.~\ref{fig:parameter}, it can be further found that DCC has produced a very small mAP value especially for a trial performed on T$\rightarrow$I task (128 bits), while inducing a larger fluctuation on different trials. That is, the mAP values corresponding to the maximum-minimum (Max-Min) difference  and standard deviation are a bit large. The main reason lies in that DCC optimization is an approximate solution and may fall into a local minimum during the learning process, which may produce unstable retrieval performances. In contrast, the proposed E-RCD algorithm can not only yield very competitive performance in various retrieval tasks, but also achieve a relatively stable retrieval performance. The average mAP values derived from ten trials do not change significantly, whereby the values of max-min difference and standard deviation are always lower than the results generated by the DCC optimization. The experimental results consistently validate the advantage of the proposed E-RCD scheme in discrete optimization,
and the proposed MTFH learning framework is beneficial to produce more effective and stable hash codes.

\begin{table}[!t]
\caption{  Results (mAP) of different optimization schemes on Wiki dataset.}\setlength{\tabcolsep}{0.22cm}
\vspace{-0.4cm}
\begin{center}
\begin{tabular}{|c|c|c|c|c|}
\hline
	\multicolumn{2}{|c|}{\multirow{2}*{Task (bits)}} & \multicolumn{1}{c|}{average mAP} & \multicolumn{1}{c|}{max-min value} & \multicolumn{1}{c|}{standard deviation} \cr \cline{3-5}
	\multicolumn{2}{|c|}{} & DCC/E-RCD & DCC/E-RCD & DCC/E-RCD \cr \cline{3-5}
	\hline
	\multirow{2}{*}{I$\rightarrow$T}
		& 32 & 0.3379/\textbf{0.3555} & 0.0526/\textbf{0.0248} & 0.0163/\textbf{0.0066} \\
		& 128 & 0.3342/\textbf{0.3418} & 0.0420/\textbf{0.0227} & 0.0143/\textbf{0.0068}  \\
		\hline
	\multirow{2}{*}{T$\rightarrow$I}
		& 32 & 0.7141/\textbf{0.7171}& 0.0557/\textbf{0.0274} & 0.0163/\textbf{0.0073} \\
		& 128 & 0.7284/\textbf{0.7372} & 0.0741/\textbf{0.0271} & 0.0218/\textbf{0.0071}  \\
		\hline
\end{tabular}
\end{center}
\label{tab:optimization}
\end{table}

\subsection{Parameter Sensitivity Analysis}
There are three main parameters involved in MTFH learning framework, \emph{i.e.}, $\alpha$,  $\lambda$ and $\beta$. Specifically, $\alpha$ balances two learning items in Eq.~\eqref{eq:origin}. A larger $\alpha$ may emphasize more on hash code learning ($q_1$ length) of modality $\mathbf{X}$, and conversely ($q_2$ length) of $\mathbf{Y}$. Since our work aims to achieve cross-modal retrieval,  it is natural to set $\alpha{=}0.5$ for balancing two modalities.  As indicated in~\cite{xu2017matrix}, $\lambda$ is insensitive to the least square optimization, and it is  set at 0.1 in most cases. $\beta$ controls the learning influence, and  we further report the
performance of changing $\beta$ while fixing $\alpha$ and $\lambda$. That is, several different values, $\beta{=}\{0.0001, 0.001, 0.01, 0.1, 1\}$, are tested on benchmark datasets (MIRFlickr and NUS-WIDE-100k). The cross-modal retrieval performances tested with different $\beta$ values and obtained by MTFH\_rnd  are shown in Fig.~\ref{fig:parameter}, it can be seen that the different settings of $\beta$ just induce a minor fluctuation on the retrieval performance, and yield very stable retrieval performance on different retrieval tasks. Therefore, $\beta$ is also insensitive to the cross-modal retrieval performance.

Further, similar to SePH~\cite{lin2015semantics},  we further sample different training sizes and utilize the learnt hash functions to generate the hash codes for all instances in training dataset. Typical examples tested on NUS-WIDE-100k dataset are shown in Fig.~\ref{fig:parameter}, it can be found that the proposed MTFH method requires a bit larger training set (around 10k for I$\rightarrow$T  and 30k for T$\rightarrow$I) to produce promising results (better than SePH). Fortunately, the mAP scores obtained by MTFH increase consistently as the training set grows from 200 to 50k, but which tend to converge when the training set is larger than 60k. Comparing with SePH~\cite{lin2015semantics}, the proposed MTFH method is computationally more efficient for very large-scale datasets and can be adapted to various  cross-modal retrieval tasks, including paired or unpaired multi-modal data collections, in either equal or varying hash length encoding scenarios.

\subsection{Discussion and Analysis}\label{extension}
The computational complexity of the proposed MTFH framework mainly accumulates from the matrix multiplications, which can be parallelized with modern computing techniques.  In practice, the size of database may be so large that it is generally impossible to learn hash functions on the whole database, mainly due to the limitation of computational resource. One solution to such problem is to learn the hash functions on a smaller training set and extend it to out-of-sample instances~\cite{lin2015semantics}.  Although the proposed MTFH method requires a semantic correlation matrix to perform the retrieval task, the  multiplication of a small matrix is very easy to implement and the retrieval time has no substantial changes. As shown in Table~\ref{tab:retrievaltime}, if the equal hash length setting is employed, the retrieval times (averaged in five runs)  of 100 queries obtained by SePH~\cite{lin2015semantics}, GSePH~\cite{mandal2017generalized} and MTFH are within the same range. It is noted that the proposed MTFH method even reduces the retrieval time when the hash length of one modality is fairly short. For instance, the hash codes derived from the I-128\&T-16 have significantly reduced the retrieval time of I-128\&T-128 in I$\rightarrow$T task, because the shortened hash codes require less processing in kernel logistic regression and the mapping from 128 bits to 16 bits can greatly reduce the similarity calculations in retrieval process.

Further, the shortened hash codes would reduce the amount of storage memory. With the similar retrieval performance, the  competing methods require $2q_1$ bits
to store the paired training instances, while the proposed MTFH method only needs $q_1{+}q_2$ ($q_2{<}q_1$) bits to store such paired instances. For instance, if the number $n$ of training pairs is very large, performance of I-32\&T-128 is comparable to the result produced by I-128\&T-128, but with significantly reduced storage space, \emph{i.e.}, $96n{-}320$ bits, $\{\mathbf{H}_1,\mathbf{H}_2\}{\in} \mathbb{R}^{32\times 128}$. Taking  the larger NUS-WIDE-All dataset for example, the best I$\rightarrow$T and T$\rightarrow$I  retrieval performances obtained by the baseline methods are generated by SePH\_km  with hash pair I-128\&T-128, as shown in Table~\ref{tab:map}. In contrast to this, the proposed MTFH approach with  hash pair I-32\&T-128 has yielded the improved retrieval performances over SePH\_km, while saving the storage space of around 17M (million) bits. Therefore, the proposed MTFH method is able to store a smaller number of bits when there exist a large number of multi-modal dataset.

Also, we evaluate the retrieval performances under the same memory budget (the storage memory of correlation matrix is ignored due to its very small size). Representative results are shown in Table~\ref{tab:retrievaltime}, it shows that the proposed MTFH with hash pairs I-48\&T-80, I-80\&T-48, I-32\&T-96 and I-96\&T-32 have yielded the better retrieval performance than that generated by hash pair I-64\&T-64 in SePH~\cite{lin2015semantics} and GSePH~\cite{mandal2017generalized}, while in some cases these varying hash encoding schemes  produce improved retrieval performance over equal hash length encoding scenario. For instance, the hash pair I-48\&T-80 has delivered the largest mAP score on I$\rightarrow$T task, when tested on MIRFlickr dataset. Therefore, the proposed MTFH framework is flexible enough to facilitate different retrieval tasks. It is pointed out that the unequal hash length encoding of multi-modal data may produce better cross-modal retrieval performance with appropriate length selection, otherwise it may also bring the negative effect to the retrieval performance. For instance, in case of I$\rightarrow$T task on the Wiki dataset, it can be found that the hash pair I-128\&T-16 shows the poor retrieval performance in comparison with the pair I-16\&T-16. The main reason lies that the unequal hash length encoding with significantly different bits may degrade the discriminative power of mapping codes, which  subsequently degrade the retrieval performance. Therefore, the appropriate length selection in varying hash length encoding scheme is necessary  for heterogeneous data representation.

\begin{table}[!t]
\caption{The retrieval time  tested on 100 queries (seconds averaged in five runs), and mAP scores recorded under similar memory budget.}\setlength{\tabcolsep}{0.07cm}
\vspace{-0.4cm}
\begin{center}
\begin{tabular}{|c|c|c|p{1cm}<{\centering}|p{1cm}<{\centering}|p{1cm}<{\centering}|p{1cm}<{\centering}|}
\hline
\multicolumn{ 1}{|c|}{\multirow{2}*{Metric}}& \multicolumn{ 1}{c|}{\multirow{2}*{Method}} & \multicolumn{ 1}{c|}{\multirow{2}*{Bit length}} & \multicolumn{ 2}{c|}{{ WIKI}} & \multicolumn{ 2}{c|}{{MIRFlickr}} \\
\cline{4-7}
\multicolumn{ 1}{|c|}{} & \multicolumn{ 1}{c|}{} & \multicolumn{ 1}{c|}{} &  {I$\rightarrow$T} &  {T$\rightarrow$I} &  {I$\rightarrow$T} &  {T$\rightarrow$I} \\
\hline
\multirow{8}{*}{\makecell[cc]{Retrieval \\ time \\(second)}}  & \multirow{2}{*} {SePH~\cite{lin2015semantics}} & {I-16\&T-16} &    0.0335  &    \textbf{0.0279}  &    0.1849  &    0.1867 \\
 & & {I-128\&T-128} & { 0.0587 } & { 0.0583 } & { 0.3997 } & { 0.4013 }\\
\cline{2-7}
 & \multirow{2}{*} {GSePH~\cite{mandal2017generalized}} & {I-16\&T-16} &    \textbf{0.0334}  &    {0.0283}  &    \textbf{0.1805}  &    \textbf{0.1843}\\
 & & {I-128\&T-128} & { 0.0585 } & { 0.0592 } & { 0.4083 } & { 0.4075 }\\
\cline{2-7}
 & \multirow{4}{*} {MTFH} &  {I-16\&T-16} &    0.0340  &    0.0295  &    0.1877 &    0.1916\\
 & & {I-128\&T-128} & { 0.0611 } & { 0.0608 } & {0.4148} & { 0.4115 }\\
 & & {I-16\&T-128} &    0.0597  &    0.0299  &    0.4103  &    0.1923 \\
 & & {I-128\&T-16} &    0.0345  &    0.0588  &    0.1914  &    0.4087 \\
 \hline
  \hline
 \multirow{5}{*}{\makecell[cc]{Retrieval\\ result \\(mAP)}}  & {SePH~\cite{lin2015semantics}} & {I-64\&T-64} &    0.3135  &    0.6570  &    0.6799  &    0.7374 \\
\cline{2-7}
 &  {GSePH~\cite{mandal2017generalized}} & {I-64\&T-64} &    0.3101  &    0.6679  &    0.6768  &    0.7313\\
\cline{2-7}
 & \multirow{5}{*} {MTFH} &  {I-64\&T-64} &   0.3454 &     0.7365 &     0.7592 &     0.8198\\
 & & {I-32\&T-96} & 0.3572   &  0.7339   &      0.7674 &     0.8213\\
 & & {I-96\&T-32} & \textbf{0.3588}  &  0.7342   &     0.7613  &     0.8186 \\
  & & {I-48\&T-80} & 0.3416 &     \textbf{0.7370} &      \textbf{0.7680} &     0.8224\\
 & & {I-80\&T-48} &  0.3390 &     0.7199 &     0.7612  &     \textbf{0.8280} \\
 \hline
 \end{tabular}
\end{center}
\label{tab:retrievaltime}
\end{table}

Besides, we notice that  the varying hash codes of different lengths can be generated by separately training two hash functions for each modality. However,  on the one hand, the varying hash codes learned in a separate way naturally weakens the connection within the same modality and often fails to preserve the semantic similarity between the heterogeneous samples due to the accumulated error. On the other hand, the hash codes of different lengths learned separately cannot  be compared directly. In contrast to this, the proposed  MTFH framework exploits an efficient objective function to jointly learn the modality-specific hash codes with different lengths, while simultaneously excavating two semantic correlation matrices to ensure heterogeneous data comparable.

It is observed form the experimental results that the proposed MTFH framework can well generalize and facilitate cross-modal retrieval in various challenging scenarios, and the  merits of using unequal hash codes are three-fold: 1) The utilization of unequal hash codes can adapt to an even more challenging cross-modal retrieval scenario, \emph{i.e.}, the hash representations from heterogeneous modalities are stored by different code lengths in the database; 2) It is beyond the limitations of equalized hash length representation of multi-modal data, by allowing varying hash length encoding for different data modalities; 3) It often produces the improved retrieval performance under same memory budget, while the shorten hash codes could reduce the storage space under similar retrieval performance. It should be noted that most extensions to multiple modalities either select the paired multi-modal data for training or employ the unified hash code for heterogeneous data representation, \emph{e.g.}, CMFH~\cite{ding2016large} and SMFH~\cite{tang2016supervised}. Specifically, the semantic affinity matrix with embedding supervision  is constructed only from two modalities~\cite{lin2015semantics,mandal2017generalized}. If the data samples from heterogeneous modalities are paired, the related works can be extended to three or more modalities, \emph{e.g.}, SePH\cite{lin2015semantics}, otherwise it is impractical to project unpaired data into a common semantic space and utilize a unified hash code to represent each data point, \emph{e.g.}, GSePH~\cite{mandal2017generalized}. The proposed MTFH is, by design, a flexible cross-modal hashing framework to handle both paired and unpaired multi-modal data collections, in either equal or varying hash length settings. Evidently, the proposed MTFH approach is able to handle all retrieval tasks reported in GSePH, while adapting to unequal hash length encoding scenario.   Remarkably, if the to-be-learnt code lengths of heterogeneous modalities are different, it is impractical to unify them in a common representation. In the current form, the proposed framework has the bottleneck for extension to more modalities and we will study it in  future work.

\section{Conclusion}\label{conclusion}
This paper has proposed a generalized and flexible Matrix Tri-Factorization Hashing (MTFH) framework for efficient cross-modal retrieval, which can seamlessly work in various challenging tasks including paired or unpaired multi-modal data, and equal or varying hash length encoding scenarios. More specifically, MTFH exploits an efficient objective function to jointly learn the  modality-specific hash codes with different length settings, while simultaneously learning two semantic correlation matrices to correlate the semantic consistency between two modalities and ensure the heterogeneous data comparable. Meanwhile, an efficient discrete optimization algorithm is presented for MTFH without relaxation such that the learned hash codes are more effective to preserve the semantic structure of multi-modal data. As a result, the derived hash codes are more semantically meaningful than those generated by traditional matrix hashing methods. To the best of our knowledge, this work is the first attempt to learn varying hash codes of different lengths for  heterogeneous data comparable and efficient cross-modal retrieval. Extensive experiments on various retrieval tasks have verified its outstanding performance. Our future work will focus on exploiting the optimum hash length with respect to each modality to carry out cross-modal retrieval task, as well as the adaptivity on a small training dataset and the extensions to more modalities.

\ifCLASSOPTIONcaptionsoff
  \newpage
\fi

\vspace{-1.2cm}
\begin{IEEEbiography}[{\includegraphics[width=1in,height=1.25in,clip,keepaspectratio]{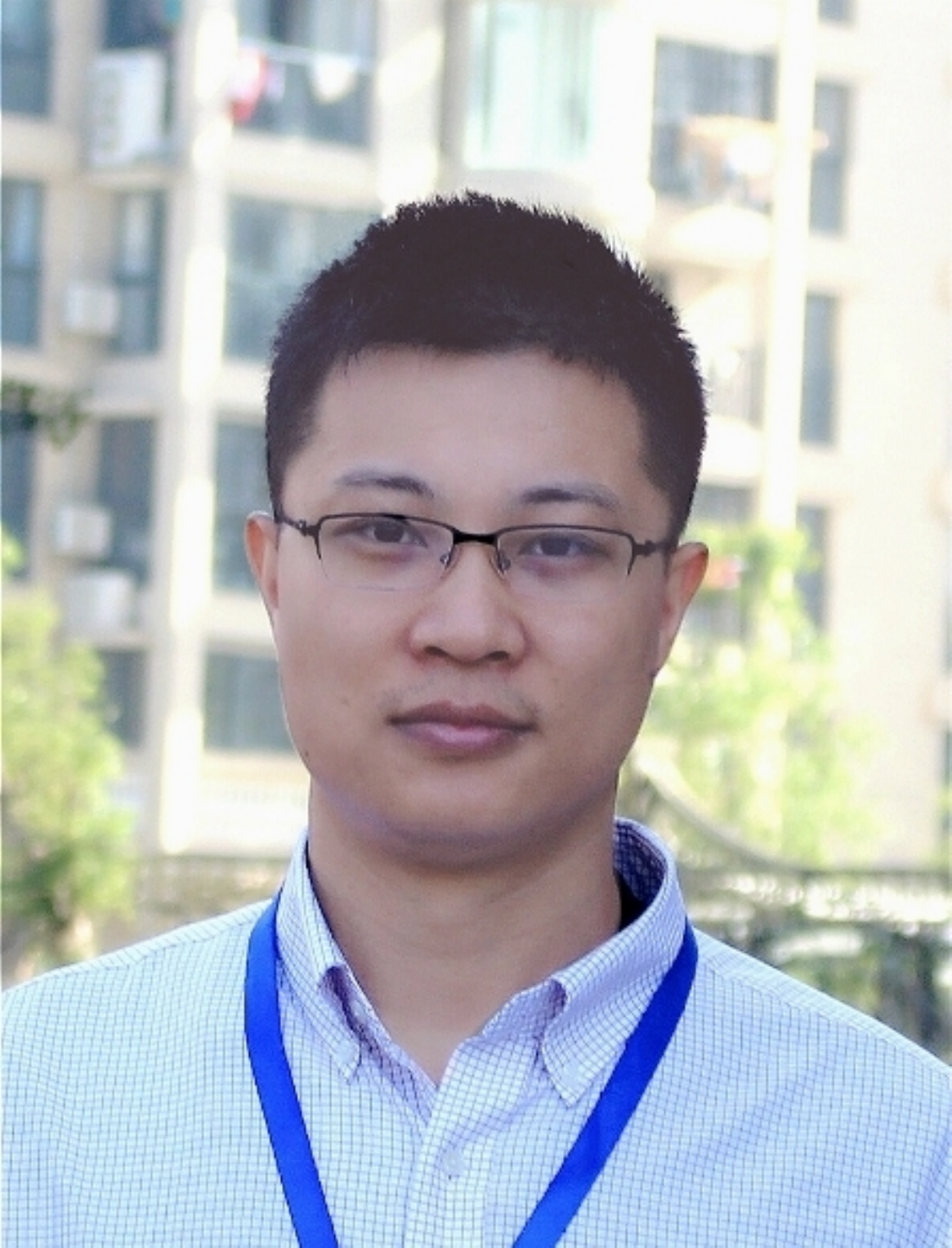}}]{Xin Liu}
received the Ph.D. degree in computer
science from Hong Kong Baptist University, Hong Kong, in 2013. He was a visiting scholar with  Computer \& Information Sciences Department, Temple University, Philadelphia, USA, from 2017 to 2018. Currently, he is
an Associate Professor with the Department of Computer Science and Technology, Huaqiao University, Xiamen, China, and also a Research Fellow with the State Key Laboratory of Integrated
Services Networks, Xidian University, Xi'an, China.  His present research interests include multimedia analysis, computer vision, pattern recognition and machine learning.  He is a member of the IEEE.
\end{IEEEbiography}
\vspace{-1.1cm}
\begin{IEEEbiography}[{\includegraphics[width=1in,height=1.25in,clip,keepaspectratio]{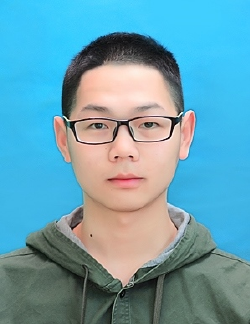}}]{Zhikai Hu}
received his B.S. degree in computer
science from China Jiliang University, Hangzhou, China, in 2015, and the M.S. degree
in computer science from  Huaqiao University, Xiamen, China, in 2019. He is currently a Research Assistant with the Department of Computer Science,
Hong Kong Baptist University. His present research interests include  information retrieval, pattern recognition and data mining. He is a student member of the IEEE.
\end{IEEEbiography}
\vspace{-1.2cm}
\begin{IEEEbiography}[{\includegraphics[width=1in,height=1.25in,clip,keepaspectratio]{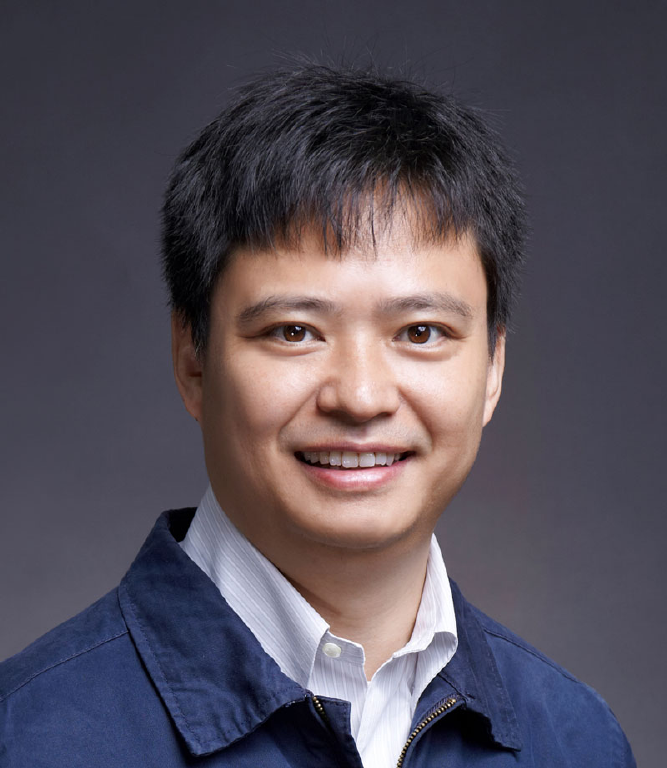}}]{Haibin Ling}
 received the B.S. and M.S. degrees from Peking University in 1997 and 2000, respectively, and the Ph.D. degree from the University of Maryland, College Park, in 2006. From 2000 to 2001, he was an assistant researcher at Microsoft Research Asia. From 2006 to 2007, he worked as a postdoctoral scientist at the University of California Los Angeles. In 2007, he joined Siemens Corporate Research as a research scientist. From 2008 to 2019, he worked as a faculty member of the Department of Computer Sciences at Temple University. In fall 2019, he joined the Computer Science Department of Stony Brook University where he is now a SUNY Empire Innovation Professor. His research interests include computer vision, augmented reality, medical image analysis, and human computer interaction. He received Best Student Paper Award at ACM UIST in 2003, and NSF CAREER Award in 2014. He serves as Associate Editors for several journals including IEEE Trans. on Pattern Analysis and Machine Intelligence (PAMI), Pattern Recognition (PR), and Computer Vision and Image Understanding (CVIU). He has served or will serve as Area Chairs for CVPR 2014, 2016, 2019 and 2020.
\end{IEEEbiography}
\vspace{-1.1cm}
\begin{IEEEbiography}[{\includegraphics[width=1in,height=1.25in,clip,keepaspectratio]{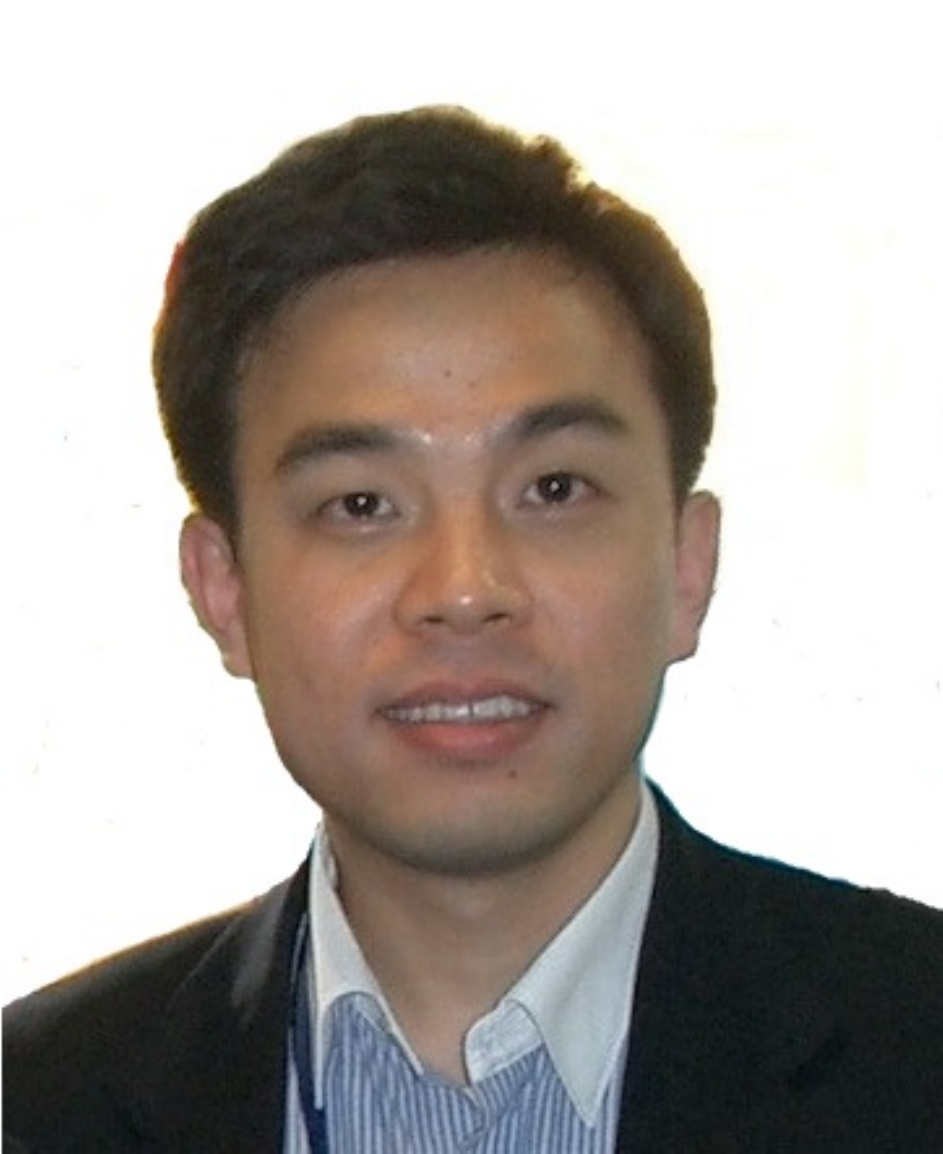}}]{Yiu-ming Cheung}
received his Ph.D. degree from the Department of Computer
Science and Engineering, Chinese University of
Hong Kong, Hong Kong. He is currently a Full Professor with the
Department of Computer Science, Hong Kong
Baptist University, Hong Kong. His current
research interests include machine learning, pattern
recognition and visual computing.
Prof. Cheung is the Founding Chairman of the Computational Intelligence Chapter of the
IEEE Hong Kong Section. He serves as an Associate Editor for the IEEE Transactions on Neural Networks and Learning Systems, Pattern Recognition,
 Knowledge and Information Systems, and the International Journal of
Pattern Recognition and Artificial Intelligence. He is an IEEE Fellow, IET/IEE Fellow and BCS Fellow.
\end{IEEEbiography}





\end{document}